\documentclass{amsart}

\usepackage{times}
\usepackage{algorithmic}
\usepackage{algorithm}
\usepackage{graphicx}
\usepackage{subfigure}
\usepackage{bbm}
\usepackage{amsmath, amsthm, amssymb}
\usepackage{color}
\usepackage[toc,page]{appendix}

\def\etal{{\em et al.\/}\, }

\usepackage{amsmath, amsthm, amssymb}
\newtheorem{proposition}{Proposition}
\newtheorem{corollary}{Corollary}
\newtheorem{theorem}{Theorem}
\newtheorem{lemma}{Lemma}

\title{A Feature Selection Method for Multivariate Performance Measures}
\author[Q. Mao]{Qi Mao}
\author[Ivor W. Tsang]{Ivor W. Tsang}
\thanks{Qi Mao and Ivor W. Tsang are with School of Computer Engineering,
Nanyang Technological University, Singapore 639798, e-mail \{QMAO1,IvorTsang\}@ntu.edu.sg}

\begin{document}
\maketitle

\begin{abstract}
Feature selection with specific multivariate performance measures is the key to the success of many applications, such as image retrieval and text classification. The existing feature selection methods are usually designed for classification error.
In this paper, we propose a generalized sparse regularizer. Based on the proposed regularizer, we present a unified feature selection framework for general loss functions. In particular, we study the novel feature selection paradigm by optimizing multivariate performance measures. The resultant formulation is a challenging problem for high-dimensional data. Hence, a two-layer cutting plane algorithm is proposed to solve this problem, and the convergence is presented.
In addition, we adapt the proposed method to optimize multivariate measures for multiple instance learning problems.
The analyses by comparing with the state-of-the-art feature selection methods show that the proposed method is superior to others.
Extensive experiments on large-scale and high-dimensional real world datasets show that the proposed method outperforms $l_1$-SVM  and SVM-RFE when choosing a small subset of features, and achieves significantly improved performances over SVM$^{perf}$ in terms of $F_1$-score.
\end{abstract}

\section{Introduction}\label{sec:introduction}

Machine learning methods have been widely applied to a variety of learning tasks (\emph{e.g.} classification, ranking, structure prediction, etc) arising in computer vision, text mining, natural language processing and  bioinformatics applications. Depending on applications, specific performance measures are required to evaluate the success of a learning algorithm. For instance, the error rate is a sound judgment for evaluating the classification performance of a learning method on datasets with balanced positive and negative examples.  On the contrary, in text classification where positive examples are usually very few, one can simply assign all testing examples with the negative class (the major class), this trivial solution can easily achieve very low error rate due to the extreme imbalance of the data. However, the goal of text classification is to correctly detect positive examples.
Hence, the error rate is considered as a poor criterion for the problems with highly skewed class distributions~\cite{Joachims05}.
To address this issue, $F_1$-score and Precision/Recall Breakeven Point (PRBEP) are employed as the evaluation criteria for text classification. Besides this, in information retrieval, search engine systems are required to return the top $k$ documents (images) with the highest precision because most users only scan the first few of them presented by the system, so precision/recall at $k$ are preferred choices.

Instead of optimizing the error rate, Support Vector Machine for multivariate performance measures ($\textrm{SVM}^{perf}$) \cite{Joachims05} was proposed to directly optimize the losses based on a variety of multivariate performance measures. A smoothing version of $\textrm{SVM}^{perf}$ \cite{Zhang2011} was proposed to accelerate the convergence of the optimization problem specially designed for PRBEP and area under the Receiver Operating Characteristic curve (AUC). Structural SVMs are considered as the general framework for optimizing a variety of  loss functions \cite{Teo10,Joachims09,Tsochantaridis05}. Other works optimize specific multivariate performance measures, such as F-score \cite{Musicant2003}, normalize discount cumulative gain (NDCG) \cite{Valizadengan2009}, ordinal regression \cite{Joachims06}, ranking loss \cite{Le2007} and so on.

For some real applications, such as image and document retrievals, a set of sparse yet discriminative features is a necessity for rapid prediction on massive databases.  However, the learned weight vector of the aforementioned methods  is usually non-sparse.
In addition, there are many noisy or non-informative features in text documents and images.
Even though the task-specific performance measures can be optimized directly, learning with these noisy or non-informative features may still hurt both prediction performance and efficiency. To alleviate these issues, one can resort to embedded feature selection methods~\cite{embedded2006}, which can be categorized into the following two major directions.

One way is to consider the sparsity of a decision weight vector $\textbf{w}$   by replacing  $l_2$-norm $\|\textbf{w}\|_2$ regularization in the structural risk functional (\emph{e.g.} SVM, logistic regression) with  $l_1$-norm  $\|\textbf{w}\|_1$~\cite{Zhu03,Fung04,Ng04}. A thorough
study to compare several recently developed $l_1$-regularized algorithms has been conducted in~\cite{YuanLR2009}. According to this study, coordinate descent method
using one-dimensional Newton direction (CDN) achieves the
state-of-the-art performance by solving  $l_1$-regularized
models on large-scale and high-dimensional datasets. To achieve a sparser solution, the Approximation of the zeRO norm Minimization (AROM) was proposed \cite{Weston03} to optimize $l_0$ models. Its resultant problem is non-convex, so it easily suffers from local optima. However, the recent results  \cite{Liu2007} and theoretical studies \cite{Lin2010,ZhangJMLR2010} have showed that $l_p$ models (where $p<1$) even with a local optimal solution can achieve better prediction performance than convex $l_1$ models, which are asymptotically biased~\cite{Liu2007}.

Another way is to sort the weights of a SVM classifier and remove the smallest weights iteratively, which is known as SVM with Recursive Feature Elimination (SVM-RFE)~\cite{Guyou02}. However, as discussed in \cite{Xu09}, such nested ``monotonic'' feature selection scheme leads to suboptimal performance. Non-monotonic feature selection (NMMKL) \cite{Xu09} has been proposed to solve this problem, but each feature corresponding to one kernel makes NMMKL infeasible for high-dimensional problems.
Recently, Tan \etal \cite{Tan10} proposed Feature Generating Machine (FGM), which shows great scalability to non-monotonic feature selection on large-scale and very high-dimensional datasets.

The aforementioned feature selection methods~\cite{YuanLR2009,Weston03,Guyou02,Xu09,Tan10} are usually designed for optimizing classification error only. To fulfill the needs of different applications, it is imperative to have a feature selection method designed for optimizing task-specific  performance measures.

To this end, we first propose a generalized sparse regularizer for feature selection. After that, a unified feature selection framework is presented for general loss functions based on the proposed regularizer. Particularly, in this paper, optimizing multivariate performance measures is studied in this framework. To our knowledge,
this is the first work to optimize multivariate performance
measures for feature selection. Due to  exponential number of constraints brought by non-smooth multivariate loss functions~\cite{Joachims05,Joachims09} and exponential number of feature subset combinations~\cite{Tan10}, the resultant optimization problem is very challenging for high-dimensional data. To tackle this challenge, we propose a two-layer cutting plane algorithm, including \emph{group feature generation} (see Section~\ref{sec:generation}) and \emph{group feature selection} (see Section~\ref{sec:selection}), to solve this problem effectively and efficiently. Specifically, Multiple Kernel Learning (MKL) trained in the primal by cutting plane algorithm is proposed to deal with exponential size of constraints induced by multivariate losses.

This paper is an extension of our preliminary  work \cite{Mao2011}.
The main contributions of this paper are listed as follows.

\begin{itemize}
\item The implementation details and the convergence proof of the proposed two-layer cutting plane algorithm
and MKL algorithm trained in the primal are presented.

\item Connections to a variety of the state-of-the-art feature selection methods including SKM \cite{Bach04}, NMMKL \cite{Xu09}, $l_1$-SVM \cite{YuanLR2009}, $l_0$-SVM \cite{Weston03} and FGM \cite{Tan10} are discussed in details. By comparing with these methods, the advantages of our proposed methods are summarized as follows:

    \begin{enumerate}
       \! \item The tradeoff parameter $C$ in $l_1$ SVM \cite{YuanLR2009} is too sensitive to be tuned properly since it controls both margin loss and the sparsity of $\mathbf{w}$. However, our method alleviates this problem by introducing an additional parameter $B$ to control the sparsity of $\mathbf{w}$. This separation makes parameter tuning for our methods  much easier than those of SKM \cite{Bach04} and $l_1$ SVM.

        \item NMMKL \cite{Xu09} uses the similar parameter separation strategy, but it is intractable for this method to handle high-dimensional datasets, let alone optimize multivariate losses. The proposed method can readily optimize multivariate losses for high-dimensional problems.

        \item FGM \cite{Tan10} is a special case of the propose framework when optimizing square hinge loss with indicator variables in integer domain. The proposed framework is formulated in the real domain for general loss functions. In particular, we provide a natural extension of FGM for multivariate losses.

        \item The proposed framework can be interpreted by $l_0$-norm constraint, so it can be considered as one of $l_0$ methods. This gives another interpretation of the additional parameter $B$.
    \end{enumerate}

\item Recall that Multiple-Instance Learning via Embedded instance Selection (MILES) \cite{Chen06}, which transforms multiple instance learning (MIL) into a feature selection problem by embedding bags into an instance-based feature space and selecting the most important features, achieves state-of-the-art performance for multiple instance learning problems.
Under our unified feature selection framework, we extend MILES and study  MIL for multivariate performance measure. To our best knowledge, this is seldom studied in MIL scenarios, but it is important for the real world applications of MIL tasks.

\item Extensive experiments on several challenging and very high-dimensional real world datasets show that the proposed method yields better performance than the state-of-the-art feature selection methods, and outperforms SVM$^{perf}$ using all features in terms of multivariate performance measures. The experimental results on the multiple instance dataset show that our proposed method achieves promising results.
\end{itemize}

The rest of the paper is organized as follows: We briefly review $\textrm{SVM}^{perf}$ in Section \ref{sec:svm-perf}. We then introduce the proposed generalized sparse regularizer in Section \ref{sec:generalized-spase-regularizer}. In particular, we study the feature selection framework for multivariate performance measures, its algorithm and its application to multiple instance learning in Section \ref{sec:feature-selection-for-multivariate},~\ref{sec:algorithm} and \ref{sec:mil-multi-variate-measure}, respectively. Section \ref{sec:relationship} gives the analysis of connections to a variety of feature selection methods. The extensive empirical results are shown in Section \ref{sec:experimetns}. Finally, conclusive remarks are presented in the last section.

In the sequel, $\mathbf{A}\succeq{\bf 0}$ means that the matrix $\mathbf{A}$ is symmetric and positive semidefinite (psd). We denote the transpose of a vector/matrix by the superscript $^T$ and $l_p$ norm of a vector $\textbf{v}$ by $||\textbf{v}||_p$. Binary operator $\odot$ represents the elementwise product between two vectors/matrices.

\section{SVM for Multivariate Performance Measure} \label{sec:svm-perf}

Given a training sample of input-output pairs $(\textbf{x}_i, y_i) \in \mathcal{X} \times \mathcal{Y}$ for $i=1,\ldots,n$ drawn from some fixed but unknown probability distribution with $\mathcal{X} \subseteq R^m$ and $\mathcal{Y} \in \{-1,+1\}$. The learning problem is treated as a multivariate prediction problem by defining the hypotheses $\overline{h}: \overline{\mathcal{X}} \rightarrow \overline{\mathcal{Y}}$ that map a tuple $\overline{\textbf{x}} \in \overline{\mathcal{X}}$ of $n$ feature vectors $\overline{\textbf{x}}=(\textbf{x}_1,\ldots,\textbf{x}_n)$ to a tuple $\overline{y} \in \overline{\mathcal{Y}}$ of $n$ labels $\overline{y}=(y_1,\ldots,y_n)$ where $\overline{\mathcal{X}} = \mathcal{X} \times \ldots,\mathcal{X}$ and $\overline{\mathcal{Y}} \subseteq \{-1,+1\}^n$. The linear discriminative function of SVM$^{perf}$  is defined as
\begin{equation}
\overline{h}_{\textbf{w}}(\overline{\textbf{x}}) = \arg \max_{\overline{y}' \in \overline{\mathcal{Y}}} f(\overline{\textbf{x}},\overline{y}') =\arg \max_{\overline{y}' \in \overline{\mathcal{Y}}} \sum_{i=1}^n y'_i \textbf{w}^T \textbf{x}_i, \label{eq:ldf-mpm}
\end{equation}
where $\textbf{w}=[w_1,\ldots,w_m]^T$ is the weight vector.

To learn the hypothesis (\ref{eq:ldf-mpm}) from training data, large margin method is employed to obtain the good generalization performance by enforcing the constraints that the decision value of the ground truth labels $\overline{y}$ should be larger than any possible labels $\overline{y}' \in \overline{\mathcal{Y}} \setminus \{\overline{y}\}$, i.e., $f(\overline{\textbf{x}},\overline{y}') \geq f(\overline{\textbf{x}},\overline{y}') + \Delta(\overline{y},\overline{y}' )$, where $\Delta(\overline{y},\overline{y}' )$ is some type of multivariate loss functions (several instantiated losses are presented in Section \ref{sec:violate}).
Structural SVMs \cite{Tsochantaridis05,Joachims09} are proposed to solve the corresponding soft-margin case by $1$-slack variable formula as,
\begin{eqnarray}
\min_{\textbf{w},\xi \geq 0} \!\!\!\!\!&&\!\!\!\!\! \frac{1}{2} \|\textbf{w}\|^2_2 + C \xi \label{prob:svm-mpm}\\
\textrm{s.t.} \!\!\!\!\!&&\!\!\!\!\! \forall \overline{y}' \in \overline{\mathcal{Y}} \backslash \overline{y} : {\textbf{w}}^T \sum_{i=1}^n (y_i - y'_i) \textbf{x}_i \geq \Delta(\overline{y},\overline{y}' ) - \xi, \nonumber
\end{eqnarray}
where $C$ is a regularization parameter that trades off the empirical risk and the model complexity.

The optimization problem (\ref{prob:svm-mpm}) is convex, but there is the exponential size of constraints. Fortunately, this problem can be solved in polynomial time by adopting the sparse approximation algorithm of structural SVMs. As shown in \cite{Joachims05}, optimizing the learning model subject to one specific multivariate measure can really boost the performance of this measure.

\section{Generalized Sparse Regularizer} \label{sec:generalized-spase-regularizer}

In this paper, we focus on minimizing the regularized empirical loss functional as
\begin{equation}
\min_{\mathbf{w}} \Omega(\mathbf{w}) + C \ell(\mathbf{w}), \label{prob:general-function}
\end{equation}
where $\Omega(.)$ is a regularization function and $\ell(.)$ is any loss function, including multivariate performance measure losses.

Since $l_2$-norm regularization is used in (\ref{prob:svm-mpm}), the learned weight vector $\textbf{w}$ is non-sparse, and so the linear discriminant function in (\ref{eq:ldf-mpm}) would involve many features for the prediction.
As discussed in Section \ref{sec:introduction}, selecting a small set of discriminative features is crucial to many real applications.
In order to enforce the sparsity on $\textbf{w}$, we propose a new sparse regularizer \[\Omega(\mathbf{w}) = \min_{\mathbf{d} \in \mathcal{D}}\frac{1}{2}\sum_{j=1}^m \frac{|w_j|^p}{d_j},\] where $\mathbf{d}$ is in the real domain of $\mathcal{D}= \{ \mathbf{d} | \sum_{j=1}^m d_j = B, 0 \leq d_j \leq 1, \forall j=1,\ldots,m \}$, $p>0$ and $B>0$ are two parameters.
The optimal solution of the new proposed regularizer should satisfy $w_j = 0$ if $d_j = 0$ since $|w_j|^p = 0$ with $p>0$ induces $w_j = 0$, otherwise the objective value approaches to infinite. The $l_1$-norm constraint $\sum_{j=1}^m d_j = B $ and $0 \leq d_j \leq 1$ will force some $d_j$ to be zero, so the corresponding $w_j$ is zero, $\forall j=1,\ldots,m$. Hence, the parameter $B$ is interpreted as a budget to control the sparsity of $\mathbf{w}$.

This regularizer is similar to SimpleMKL \cite{Rakotomamonjy08} with each feature corresponding to one kernel, but SimpleMKL is a special case of $\mathcal{D}$ with $B=1$, which also can be interpreted by the quadratic variational formulation of $l_1$ norm~\cite{Bach2012}. However, it is different from $l_1$ when $B \not = 1$. To explain the difference, we consider the problem (\ref{prob:svm-mpm}) under the general framework~(\ref{prob:general-function}). In the separable case, parameter $C$ does not affect the optimum solution since the error $\xi = 0$. If $l_1$ norm is applied to replace $l_2$ in Problem (\ref{prob:svm-mpm}), the sparsity of $\mathbf{w}$ will be fixed once optimal solution is reached.
Hence, parameter $B$ in $\mathcal{D}$ now can be considered as the only factor to enforce sparsity on $\mathbf{w}$. However, in the non-separable case where errors are allowed, parameter $C$ will also influence the sparsity of $\mathbf{w}$, but $B$ is expected to enforce the sparsity of $\mathbf{w}$ more explicitly
when $C$ becomes larger. This argument will be empirically justified in Section \ref{sec:parameter-sensitive-analysis}.

The learning algorithm with the proposed generalized sparse regularizer is formulated as
\begin{eqnarray}
\min_{\mathbf{d} \in \mathcal{D}} \min_{\textbf{w}} \!\!\!\!\!&&\!\!\!\!\!\frac{1}{2}\sum_{j=1}^m \frac{|w_j|^p}{d_j} + C \ell(\mathbf{w}). \label{prob:svm-simplemkl-framework}
\end{eqnarray}
This formulation is more general for feature selection.

\begin{lemma}
If $p \geq 2$, Problem (\ref{prob:svm-simplemkl-framework}) is jointly convex with respect to  $\textbf{w}$ and $\mathbf{d}$; otherwise, it is not jointly convex.
\end{lemma}
\begin{proof}
We only need to prove that, if $p \geq 2$, $g(w_j,d_j)=\frac{|w_j|^p}{d_j}$ where $d_j>0$ is  jointly convex with respect to $w_j$ and $d_j$. The convexity of $g$ in its domain is established when the following holds:
$ \nabla^2 g = \left[
\begin{array}{cc}
\frac{2|w_j|^p}{d_j^3} & -\frac{p |w_j|^{p-1}}{d_j^2} \\
-\frac{p|w_j|^{p-1}}{d_j^2} & \frac{p(p-1)|w_j|^{p-2}}{d_j}
\end{array}\right]\succeq {\bf 0}
\Leftrightarrow \left[
\begin{array}{cc}
2|w_j|^2 & -p |w_j|d_j \\
-p |w_j|d_j & p(p-1)d_j^2
\end{array}\right]\succeq {\bf 0},
$
which is equivalent to  ${\bf v}^T \nabla^2 g {\bf v} \geq 0$  for any nonzero vector ${\bf v}$. \emph{WLOG}, we assume ${\bf v}=[1\; a]^T$ where $a$ is any real number, then this condition is reduced to:
$2 |w_j|^2 -2a p |w_j|d_j + a^2 p(p-1)d_j^2 \geq 0
\Leftrightarrow 2 \Big(|w_j|-\frac{a p d_j}{2} \Big)^2  \geq \frac{a^2d_j^2p(2-p)}{2}.$
This condition always holds when $p\geq 2$, which completes the proof.
\end{proof}

 In what follows, we focus on the convex formulation with $p=2$. In Section \ref{sec:relationship}, we will discuss the relationships with a variety of the state-of-the-art feature selection methods.

\section{Feature Selection for Multivariate Performance Measures} \label{sec:feature-selection-for-multivariate}

To optimize the multivariate loss functions and learn a sparse feature representation simultaneously, we propose to solve the following jointly convex problem over $\mathbf{d}$ and $(\mathbf{w}, \xi)$ in the case of $p=2$,
\begin{small}
\begin{eqnarray}
\min_{\mathbf{d} \in \mathcal{D}} \min_{\textbf{w},\xi \geq 0} \!\!\!\!\!&&\!\!\!\!\! \frac{1}{2}\sum_{j=1}^m  \frac{|w_j|^2}{d_j} + C \xi \label{prob:svm-simplemkl-general}\\
\textrm{s.t.} \!\!\!\!\!&&\!\!\!\!\! \forall \overline{y}' \in \overline{\mathcal{Y}} \backslash \overline{y} : {\textbf{w}}^T \frac{1}{n} \sum_{i=1}^n (y_i - y'_i) \textbf{x}_i \geq \Delta(\overline{y},\overline{y}' ) - \xi. \nonumber
\end{eqnarray}
\end{small}\noindent
The partial dual with respect to $(\mathbf{w},\xi)$ is obtained by Lagrangian function $\mathcal{L}(\mathbf{w},\xi,\alpha,\tau)$ with dual variables $\alpha \geq 0$ and $\tau \geq 0$ as follows:
$\frac{1}{2}\sum_{j=1}^m  \frac{|w_j|^2}{d_j} + C \xi - \tau \xi
- \sum_{\overline{y}' \in \overline{\mathcal{Y}} \backslash \overline{y}} \alpha_{\overline{y}'} ( {\textbf{w}}^T \frac{1}{n} \sum_{i=1}^n (y_i - y'_i) \textbf{x}_i - \Delta(\overline{y},\overline{y}' ) + \xi ).$
As the gradients of Lagrangian function with respect to $(\mathbf{w},\xi)$ vanish at the optimal points, we obtain the KKT conditions:
$w_j = d_j \sum_{\overline{y}' \in \overline{\mathcal{Y}} \backslash \overline{y}} \alpha_{\overline{y}'} \frac{1}{n} \sum_{i=1}^n (y_i - y'_i) x_{j,i}
\mbox{ and }
\sum_{\overline{y}' \in \overline{\mathcal{Y}} \backslash \overline{y}} \alpha_{\overline{y}'} \leq C.$
By substituting KKT conditions back to $\mathcal{L}(\mathbf{w},\xi,\alpha,\tau)$, we obtain the dual problem as
\begin{eqnarray}
\min_{\mathbf{d} \in \mathcal{D}} \max_{\alpha \in \mathcal{A}} -\frac{1}{2} \sum_{\overline{y}'} \sum_{\overline{y}''}  \alpha_{\overline{y}'}  \alpha_{\overline{y}''} Q_{\overline{y}',\overline{y}''}^{\textbf{d}} +  \sum_{\overline{y}'} \alpha_{\overline{y}'} b_{\overline{y}'}, \label{prob:our-dual}
\end{eqnarray}
where $\Delta(\overline{y},\overline{y} )=0$, $\Delta(\overline{y},\overline{y}' ) >0 $ if $\overline{y} \not = \overline{y}'$,
\begin{eqnarray*}
Q_{\overline{y}',\overline{y}''}^{\textbf{d}} \!\!\!\!\!&=&\!\!\!\!\! \sum_{j=1}^m d_j \bigg( \sum_{\overline{y}' \in \overline{\mathcal{Y}} \backslash \overline{y}} \alpha_{\overline{y}'} \frac{1}{n} \sum_{i=1}^n (y_i - y'_i) x_{j,i} \bigg)^2 \\
\!\!\!\!\!&=&\!\!\!\!\! \sum_{j=1}^m \bigg( \sum_{\overline{y}' \in \overline{\mathcal{Y}} \backslash \overline{y}} \alpha_{\overline{y}'} \frac{1}{n} \sum_{i=1}^n (y_i - y'_i) x_{j,i} \sqrt{d_j} \bigg)^2 \\
\!\!\!\!\!&=&\!\!\!\!\! \langle \textbf{a}_{\overline{y}'}, \textbf{a}_{\overline{y}''} \rangle,
\end{eqnarray*}
$\textbf{a}_{\overline{y}'} = \frac{1}{n} \sum_{i=1}^n (y_i - y'_i) (\textbf{x}_i \odot \sqrt{\textbf{d}} )$, $b_{\overline{y}'} = \frac{1}{n} \Delta(\overline{y},\overline{y}')$, and $\mathcal{A} = \{ \alpha | \sum_{\overline{y}'} \alpha_{\overline{y}'} \leq C, \alpha \geq 0 \}$. Problem (\ref{prob:our-dual}) is a challenging problem because of the exponential size of $\alpha$ and high-dimensional vector $\mathbf{d}$ for high-dimensional problems.

\section{Two-Layer Cutting Plane Algorithm}\label{sec:algorithm}
In this section, we propose a two-layer cutting plane algorithm to solve Problem (\ref{prob:our-dual}) efficiently and effectively. The two layers, namely
group feature generation  and group feature selection, will be described in Section~\ref{sec:generation} and~\ref{sec:selection}, respectively. The two-layer cutting plane algorithm will be presented in Section~\ref{sec:cutting} and~\ref{sec:violate}.

\subsection{Group Feature Generation} \label{sec:generation}

By denoting $S(\alpha,\textbf{d}) = -\frac{1}{2} \sum_{\overline{y}'} \sum_{\overline{y}''}  \alpha_{\overline{y}'}  \alpha_{\overline{y}''} Q_{\overline{y}',\overline{y}''}^{\textbf{d}} +  \sum_{\overline{y}'} \alpha_{\overline{y}'} b_{\overline{y}'}$, Problem (\ref{prob:our-dual}) turns out to be
\begin{displaymath}
\min_{\textbf{d} \in \mathcal{D}} \max_{\alpha \in \mathcal{A}} S(\alpha,\textbf{d}).
\end{displaymath}
Since domains $\mathcal{D}$ and $\mathcal{A}$ are nonempty, the function $S(\alpha^*,\mathbf{d})$ is closed and convex for all $\mathbf{d} \in \mathcal{D}$ given any $\alpha^* \in \mathcal{A}$, and the function $S(\alpha,\mathbf{d}^*)$ is closed and concave for all $\alpha \in \mathcal{A}$ given any $\mathbf{d}^* \in \mathcal{D}$, the saddle-point property:
$\min_{\textbf{d} \in \mathcal{D}} \max_{\alpha \in \mathcal{A}} S(\alpha,\textbf{d}) = \max_{\alpha \in \mathcal{A}} \min_{\textbf{d} \in \mathcal{D}}  S(\alpha,\textbf{d})$
holds \cite{Analysis2000}.

We further denote $\mathcal{F}_\textbf{d}(\alpha) = -S(\alpha,\textbf{d})$, and then the equivalent optimization problems are obtained as
\begin{eqnarray}\label{eq:minimax}
\min_{\alpha \in \mathcal{A}} \max_{\textbf{d} \in \mathcal{D}}\mathcal{F}_\textbf{d}(\alpha) \!&\!\!\mbox{or}\!\!&\! \min_{\alpha \in \mathcal{A},\gamma}\gamma : \gamma\geq \mathcal{F}_\textbf{d}(\alpha),\; \forall\textbf{d} \in \mathcal{D}.
\end{eqnarray}
Cutting plane algorithm \cite{Kelley(60)} could be used here to solve this problem.
Since $\max_{\textbf{d} \in \mathcal{D}} \mathcal{F}_\textbf{d}(\alpha) \geq \mathcal{F}_{\textbf{d}^t}(\alpha), \forall d^t \in \mathcal{D}$, the lower bound approximation of (\ref{eq:minimax}) can be obtained by
$\max_{\textbf{d} \in \mathcal{D}} \mathcal{F}_\textbf{d}(\alpha) \geq \max_{t=1,\ldots,T} \mathcal{F}_{\textbf{d}^t}(\alpha)$.
Then we minimize Problem (\ref{eq:minimax}) over the set $\{\mathbf{d}^t \}_{t=1}^T$ by,
\begin{eqnarray}
\min_{\alpha \in \mathcal{A}} \!\max_{t=1,\ldots,T} \!\mathcal{F}_{\textbf{d}^t}(\alpha) \!\!&\!\!\mbox{or}\!\!\!&\!\!\min_{\alpha \in \mathcal{A},\gamma} \!\gamma
\!:\! \gamma \!\geq\! \mathcal{F}_{\textbf{d}^t}(\alpha), \forall t\!=\!1\!,\!\ldots\!,\! T\!\!.\;\;\;  \label{prob:svm-mpm-ours_dual}
\end{eqnarray}
As from \cite{Mutapcic09}, such cutting plane algorithm can converge to a robust optimal solution within tens of iterations with the exact worst-case analysis. Specifically, for a fixed $\alpha^t$, the worst-case analysis can be done by solving,
\begin{equation}
\textbf{d}^t = \arg \max_{\textbf{d} \in \mathcal{D}} \mathcal{F}_{\textbf{d}}(\alpha^t), \label{prob:find_most_violated_d}
\end{equation}
which is referred to as the group generation procedure.
Even though Problem  (\ref{prob:svm-mpm-ours_dual}) and (\ref{prob:find_most_violated_d}) cannot be solved directly due to the exponential size of $\alpha$, we will show that they are readily solved in Section \ref{sec:selection} and Section \ref{sec:violate}, respectively.

\subsection{Group Feature Selection} \label{sec:selection}

By introducing dual variables $\mu= [\mu_1,\mu_2,\ldots,\mu_T]^T \geq 0$, we can transform (\ref{prob:svm-mpm-ours_dual})
to an MKL problem as follows,
\begin{small}
\begin{equation}
\max_{\alpha \in \mathcal{A}}\min_{\mu \in \mathcal{M}_T} -\frac{1}{2} \sum_{\overline{y}'} \sum_{\overline{y}''}  \alpha_{\overline{y}'}  \alpha_{\overline{y}''} \left( \sum_{t=1}^T \mu_t Q_{\overline{y}',\overline{y}''}^{\textbf{d}^t} \right) +  \sum_{\overline{y}'} \alpha_{\overline{y}'} b_{\overline{y}'}, \label{prob:svm-mpm-ours_dual_mu}
\end{equation}
\end{small}\noindent
where $\mathcal{M}_T = \{ \sum_{t=1}^T \mu_t = 1, \mu_t \geq 0, \forall t=1,\ldots,T\}$.

However, due to the exponential size of $\alpha$, the complexity of Problem (\ref{prob:svm-mpm-ours_dual_mu}) remains. In this case, state-of-the-art multiple kernel learning algorithms \cite{Sonnenburg06,Rakotomamonjy08,Xu08} do not work any more. The following proposition shows that we can indirectly solve Problem (\ref{prob:svm-mpm-ours_dual_mu}) in the primal form.

\begin{proposition} \label{prop:dual-primal}
The primal form of Problem (\ref{prob:svm-mpm-ours_dual_mu}) is
\begin{eqnarray}
\min_{\textbf{w}_1,\ldots,\textbf{w}_T,\xi \geq 0} \!\!\!\!\!&&\!\!\!\!\! \frac{1}{2} \bigg( \sum_{t=1}^T \|\textbf{w}_t\|_2 \bigg)^2 + C \xi \label{prob:mkl-primal}\\
\textrm{s.t.} \!\!\!\!\!&&\!\!\!\!\! \xi \geq  b_{\overline{y}'} - \sum_{t=1}^T \langle \textbf{w}_t, \textbf{a}_{\overline{y}'}^t \rangle, \forall \overline{y}' \in \overline{\mathcal{Y}} \backslash \overline{y} \nonumber.
\end{eqnarray}
According to KKT conditions, the solution of (\ref{prob:mkl-primal}) is
\begin{eqnarray}
\textbf{w}_t = \mu_t \sum_{\overline{y}'} \alpha_{\overline{y}'} \textbf{a}_{\overline{y}'}^t \label{prob:mkl-primal-solution}
\end{eqnarray}
where $\mu_t$ is a dual value of the $t^{th}$ constraint of (\ref{prob:svm-mpm-ours_dual}).
\end{proposition}
The detailed proof of Proposition \ref{prop:dual-primal} is given in the supplementary material.

Here, we define the regularization term as $\Omega(\overline{\textbf{w}}) =  \frac{1}{2} \big( \sum_{t=1}^T \|\textbf{w}_t\|_2 \big)^2$ with $\overline{\textbf{w}} = [\textbf{w}_1,\ldots,\textbf{w}_T]^T$and the empirical risk function as
\begin{equation}
R_{emp}(\overline{\textbf{w}}) = \max\bigg(0,  \max_{\overline{y}' \in \overline{\mathcal{Y}} \backslash \overline{y} } b_{\overline{y}'} - \sum_{t=1}^T \langle \textbf{w}_t, \textbf{a}_{\overline{y}'}^t \rangle  \bigg),
\end{equation}
which is a convex but non-smooth function w.r.t $\overline{\textbf{w}}$.
Then we can apply the bundle method \cite{Teo10} to solve this primal problem.
Problem (\ref{prob:mkl-primal}) is transformed as
\begin{eqnarray*}
\min_{\overline{\textbf{w}}} \mathcal{J}(\overline{\textbf{w}}) = \Omega(\overline{\textbf{w}}) + C R_{emp}(\overline{\textbf{w}}).
\end{eqnarray*}
Since $R_{emp}(\overline{\textbf{w}})$ is a convex function, its subgradient exists everywhere in its domain \cite{Hiriart-Urruty93}. Suppose $\overline{\textbf{w}}^k$ is a point where $R_{emp}(\overline{\textbf{w}})$ is finite, we can formulate the lower bound according to the definition of subgradient,
\begin{eqnarray*}
R_{emp}(\overline{\textbf{w}}) &\geq& R_{emp}(\overline{\textbf{w}}^k) + \langle \overline{\textbf{w}} - \overline{\textbf{w}}^k, \textbf{p}^k \rangle \\
&=& \langle \overline{\textbf{w}},\textbf{p}^k \rangle + R_{emp}(\overline{\textbf{w}}^k) - \langle \overline{\textbf{w}}^k, \textbf{p}^k \rangle
\end{eqnarray*}
where subgradient $\textbf{p}^k \in \partial_{\overline{\textbf{w}}} R_{emp}(\overline{\textbf{w}}^k)$ is at $\overline{\textbf{w}}^k$. In order to obtain $\textbf{p}^k$, we need to solve the following inference problem
\begin{equation}
\overline{y}^k = \arg \max_{\overline{y}' \in \overline{\mathcal{Y}} \backslash \overline{y} } b_{\overline{y}'} - \sum_{t=1}^T \langle \textbf{w}_t, \textbf{a}_{\overline{y}'}^t \rangle
\end{equation}
which is a problem of integer programming. We delay the discussion of this problem to Section \ref{sec:violate}. After that, we can obtain the subgraident $\textbf{p}^k_t = - \mathbf{a}_{\overline{y}^k}^t$, so that $R_{emp}(\overline{\textbf{w}}^k) = b_{\overline{y}^k} - \sum_{t=1}^T \langle \textbf{w}_t, \textbf{a}_{\overline{y}^k}^t \rangle = b_{\overline{y}^k} + \langle \overline{\textbf{w}}^k, \textbf{p}^k \rangle$.

Given the subgradient sequence $\textbf{p}^1,\textbf{p}^2,\ldots,\textbf{p}^K$, the tighter lower bound for $R_{emp}(\overline{\textbf{w}})$ can be reformulated as follows,
\begin{eqnarray*}
R_{emp}(\overline{\textbf{w}}) \geq R_{emp}^K(\overline{\textbf{w}})
                              = \max\Big(0, \max_{1 \leq k \leq K} \langle \overline{\textbf{w}}, \textbf{p}^k \rangle + q^k\Big),
\end{eqnarray*}
where $q^k =R_{emp}(\overline{\textbf{w}}^k) -  \langle \overline{\textbf{w}}^k, \textbf{p}^k \rangle = b_{\overline{y}^k}$. Following the bundle method \cite{Teo10}, the criterion for selecting the next point $\overline{\textbf{w}}^{K+1}$ is to solve the following  problem,
\begin{eqnarray}
\min_{\textbf{w}_1,\ldots,\textbf{w}_T,\xi \geq 0} \!\!\!\!\!&&\!\!\!\!\! \frac{1}{2} \bigg( \sum_{t=1}^T \|\textbf{w}_t\|_2 \bigg)^2 + C \xi \label{prob:mkl-reduced-primal}\\
\textrm{s.t.} \!\!\!\!\!&&\!\!\!\!\! \xi \geq  \langle \overline{\textbf{w}}, \textbf{p}^k \rangle + q^k, \forall k=1,\ldots,K.  \nonumber
\end{eqnarray}
The following Corollary shows that Problem (\ref{prob:mkl-reduced-primal})
can be easily solved by QCQP solvers, and the number of variables is independent of the number of examples.
\begin{corollary} \label{cor:small-QCQP}
In terms of Proposition \ref{prop:dual-primal}, the dual form of Problem (\ref{prob:mkl-reduced-primal}) is
\begin{eqnarray}
\max_{\alpha \in \mathcal{A}_K} \max_{\theta} \!\!\!\!\!&&\!\!\!\!\!  -\theta + \sum_{k=1}^K \alpha_k q^k \label{prob:small-QCQP}\\
\textrm{s.t.} \!\!\!\!\!&&\!\!\!\!\! \frac{1}{2}\Bigg\|\sum_{k=1}^K \alpha_k \textbf{p}_t^k\Bigg\|_2^2 \leq \theta , \forall t=1,\ldots, T \nonumber,
\end{eqnarray}
where $\mathcal{A}_K = \{\sum_{k=1}^K \alpha_k \leq C, \alpha_k \geq 0, \forall k=1,\ldots,K\}$, and which is a QCQP problem with $T+1$ constraints and $K+1$ variables.
\end{corollary}

The proof of Corollary \ref{cor:small-QCQP} follows the same derivation of Proposition \ref{prop:dual-primal} with $\textbf{p}^k_t = - \mathbf{a}_{\overline{y}^k}^t$, $q^k = b_{\overline{y}^k}$ and the size of $\alpha_k$ as $K$. Consequently, the primal variables are recovered by $\mathbf{w}_t = -\mu_t \sum_k \alpha_k \textbf{p}_t^k$.

\begin{algorithm}
   \caption{Group\_feature\_selection}
   \label{algo:group-feature-selection}
\begin{algorithmic}[1]
    \STATE Input: $\overline{\textbf{x}}=(\textbf{x}_1,\ldots,\textbf{x}_n), \overline{y}=(y_1,\ldots,y_n)$, an initial group set $\mathcal{W}$, $\epsilon$, $C$
    \STATE $\overline{\mathcal{Y}} = \emptyset$, $k = 0$
    \REPEAT
       \STATE $k = k + 1$
       \STATE Finding the most violated $\overline{y}'$
       \STATE Compute $\textbf{p}^k$ and $q^k$
       \STATE $\overline{\mathcal{Y}} = \overline{\mathcal{Y}} \cup \{ \overline{y}' \}$
       \STATE Solving Problem (\ref{prob:small-QCQP}) over $\mathcal{W}$ and $\overline{\mathcal{Y}} $
    \UNTIL{$\epsilon$-optimal}
\end{algorithmic}
\end{algorithm}
Let $\mathcal{J}_K(\overline{\textbf{w}}) = \Omega(\overline{\textbf{w}}) + C R_{emp}^K(\overline{\textbf{w}})$, the $\epsilon$-optimal condition in Algorithm \ref{algo:group-feature-selection} is $\min_{0 \leq k \leq K} \mathcal{J}(\overline{\textbf{w}}^K) - \mathcal{J}_K(\overline{\textbf{w}}^K) \leq \epsilon$. The convergence proof in \cite{Teo10} does not apply in this case as the Fenchel dual of $\Omega(\overline{\textbf{w}})$ fails to satisfy the strong convexity assumption if $K > 1$. As $K=1$, Algorithm \ref{algo:group-feature-selection} is exactly the bundle method \cite{Teo10}. When $K \geq 2$, we can adapt the proof of Theorem $5$ in \cite{Joachims09} for  the following convergence results.

\begin{theorem} \label{theorem:QCQP}
For any $0 < C, 0 < \epsilon \leq 4 R^2 C$ and any training example $(\textbf{x}_1,y_1),\ldots,(\textbf{x}_n,y_n)$, Algorithm \ref{algo:group-feature-selection} converges to the desired precision $\epsilon$ after at most,
\begin{eqnarray*}
\left\lceil \log_2 \left( \frac{\Delta}{4 R^2 C} \right) \right\rceil + \left \lceil \frac{16 R^2 C}{\epsilon} \right\rceil
\end{eqnarray*}
iterations. $R^2 = \max_{\textbf{d}^t,\overline{y}'} \|\frac{1}{n}\sum_{i=1}^n (y_i - y'_i) (\textbf{x}_i \odot \sqrt{\textbf{d}^t})\|^2$, $\Delta = \max_{\overline{y}'} \Delta(\overline{y}',\overline{y})$ and $\lceil . \rceil$ is the integer ceiling function.
\end{theorem}
\begin{proof}
We adapt the proof of Theorem 5 in \cite{Joachims09}, and sketch
the necessary changes corresponding to Problem (\ref{prob:mkl-primal}).
For a given set $\mathcal{W}_T$, the dual objective of $(\ref{prob:svm-mpm-ours_dual})$ can be reformulated as
\begin{small}
\begin{displaymath}
 \max_{\alpha \in \mathcal{A}} \min_{ \textbf{d} \in \mathcal{W}_T} \Theta_{\textbf{d}}(\alpha) = -\frac{1}{2} \sum_{\overline{y}'} \sum_{\overline{y}''}  \alpha_{\overline{y}'}  \alpha_{\overline{y}''} Q_{\overline{y}',\overline{y}''}^{\textbf{d}} +  \sum_{\overline{y}'} \alpha_{\overline{y}'} b_{\overline{y}'}.
\end{displaymath}
\end{small}\noindent
Since there are the $T$ constrained quadratic problems, we consider each $\textbf{d} \in \mathcal{W}_T$ at one time as $\max_{\alpha \in \mathcal{A}} \Theta_{\textbf{d}}(\alpha)$,
where $Q^{\textbf{d}}$ is positive semi-definite, and derivative $\partial \Theta_{\textbf{d}}(\alpha) = \textbf{b} - Q^{\textbf{d}} \alpha$. The Lemma 2 in \cite{Joachims09} states that a line search starting at $\alpha$ along an ascent direction $\eta$ with maximum step-size $C > 0$ improves the objective by at least
$\max_{0 \leq \beta \leq C}  \big\{\Theta_{\textbf{d}}(\alpha + \beta \eta) - \Theta_{\textbf{d}}(\alpha) \big\}
\geq \frac{1}{2} \min \left\{ C, \frac{\partial \Theta_{\textbf{d}}(\alpha)^T \eta}{\eta^T Q^{\textbf{d}} \eta} \right\} \partial \Theta_{\textbf{d}}(\alpha)^T \eta.$
If we consider subgradient descent method, the line search along the subgradient of objective is $\partial \Theta_{\textbf{d}^*} (\alpha)$ where $\textbf{d}^* = \min_{\textbf{d} \in \mathcal{W}_T} \Theta_{\textbf{d}}(\alpha)$. Therefore, the maximum improvement is
\begin{eqnarray}
&&\max_{0 \leq \beta \leq C}  \{\Theta_{\textbf{d}^*}(\alpha + \beta \eta) - \Theta_{\textbf{d}^*}(\alpha) \} \nonumber\\
&\geq& \frac{1}{2} \min \left\{ C, \frac{\partial \Theta_{\textbf{d}^*}(\alpha)^T \eta}{\eta^T Q^{\textbf{d}^*} \eta} \right\} \partial\Theta_{\textbf{d}^*}(\alpha)^T \eta \nonumber\\
&\geq& \frac{1}{2} \min_{\textbf{d} \in \mathcal{W}_T} \left\{ C, \frac{\partial \Theta_{\textbf{d}}(\alpha)^T \eta}{\eta^T Q^{\textbf{d}} \eta} \right\} \partial \Theta_{\textbf{d}}(\alpha)^T \eta. \label{eq:proof}
\end{eqnarray}

We can see that it is a special case of \cite{Joachims09} if $T=1$.
 According to Theorem 5 in \cite{Joachims09}, for a newly added constraint $\widehat{y}$ and some $\gamma_{\textbf{d}} > 0$, we can obtain $\partial \Theta_{\textbf{d}}(\alpha)^T \eta = \gamma_{\textbf{d}}$ by setting the ascent direction $\eta_{\widehat{y}}=1$ for the newly added $\widehat{y}$ and $\eta_{\overline{y}} = -\frac{1}{C} \alpha_{\overline{y}}$ for the others.
 Here, we  set $\gamma = \min_{\textbf{d} \in \mathcal{W}_T} \gamma_{\textbf{d}}$ so as to be the lower bound of $\partial \Theta_{\textbf{d}}(\alpha)^T \eta, \forall \textbf{d} \in \mathcal{W}_T$.
In addition, the upper bound for $\eta^T Q^{\textbf{d}} \eta \leq 4 R^2, \forall \textbf{d} \in \mathcal{W}_T$ can also be obtained by the fact that
 $\eta^T Q^{\textbf{d}} \eta = Q^{\textbf{d}}_{\widehat{y},\widehat{y}} - \frac{2}{C} \sum_{\overline{y}'} \alpha_{\overline{y}'} Q^{\textbf{d}}_{\overline{y}',\widehat{y}} + \frac{1}{C^2} \sum_{\overline{y}'} \sum_{\overline{y}''} \alpha_{\overline{y}'} \alpha_{\overline{y}''} Q^{\textbf{d}}_{\overline{y}',\overline{y}''} \leq R^2 + \frac{2}{C} CR^2 + \frac{1}{C^2} C^2 R^2 = 4 R^2, \forall \textbf{d} \in \mathcal{W}_T$.
By substituting them back to (\ref{eq:proof}), the similar result shows the increase of the objective is at least
\begin{displaymath}
\min \left\{ \frac{C \gamma}{2}, \frac{\gamma^2}{8R^2} \right\}.
\end{displaymath}

Moreover, the initial optimality gap is at most $C \Delta$. Following the remaining derivation in \cite{Joachims09}, the overall bound results are obtained.
\end{proof}

\textbf{Remark $1$:} Problem (\ref{prob:mkl-reduced-primal}) is similar to Support Kernel Machine (SKM) \cite{Bach04} in which the multiple Gaussian kernels are built on random subsets of features, with varying widths. However, our method can automatically choose the most violated subset of features as a group instead of a subset of random features. Such random features lead to a local optimum; while our method could guarantee the $\epsilon$-optimality stated in Theorem \ref{theorem:QCQP}. However, due to the extra cost of computing nonlinear kernel, the current model are only implemented for linear kernel with learned subsets of features.

\textbf{Remark $2$:} The original Problem (\ref{eq:minimax}) could be easily formulated as a QCQP problem with exponential size of variables $\alpha$ needed to be optimized and huge number of base kernels in the quadratic term. Unfortunately, the standard MKL methods cannot handle Problem (\ref{eq:minimax}) even for a small dataset, let alone the standard QCQP solver. However, Corollary \ref{cor:small-QCQP} makes it practical to solve a sequence of small QCQP problems directly using standard off-line QCQP solvers, such as Mosek. Note that state-of-the-art MKL solvers can also be used to solve the small QCQP problems, but they are not preferred because their solutions are less accurate than that of standard QCQP solvers, which can solve Problem (\ref{prob:small-QCQP}) more accurately in this case.

\subsection{The Proposed Algorithm}\label{sec:cutting}

Algorithm \ref{algo:group-feature-selection} can obtain the $\epsilon$-optimal solution for the original dual problem (\ref{prob:svm-mpm-ours_dual}). By denoting $\mathcal{G}_d(\alpha) = \frac{1}{2}||\sum_{k=1}^K \alpha_k \textbf{p}^k||_2^2 - \sum_{k=1}^K \alpha_k q^k$, the group feature generation layer can directly use the $\epsilon$-optimal solution of the objective $\mathcal{G}_{\textbf{d}} (\alpha)$ to approximate the original objective $\mathcal{F}_{\textbf{d}}(\alpha)$. The two-layer cutting plane algorithm is presented in Algorithm \ref{algo:two-layer-method}.
\begin{algorithm}
   \caption{The Two-Layer Cutting Plane Algorithm}
   \label{algo:two-layer-method}
\begin{algorithmic}[1]
    \STATE Input: $\overline{\textbf{x}}=(\textbf{x}_1,\ldots,\textbf{x}_n), \overline{y}=(y_1,\ldots,y_n)$, $\epsilon$, $C$
    \STATE $\mathcal{W} = \emptyset$, $t=0$
    \REPEAT
       \STATE $t = t + 1$
       \STATE Finding the most violated $\textbf{d}^t$
       \STATE $\mathcal{W} = \mathcal{W} \cup \{ \textbf{d}^t \}$
       \STATE Call \emph{group\_feature\_selection}($\overline{\textbf{x}}$, $\overline{y}$, $\mathcal{W}$, $\epsilon$, $C$)
    \UNTIL{$\epsilon$-optimal}
\end{algorithmic}
\end{algorithm}
From the description of Algorithm \ref{algo:two-layer-method}, it is clear to see that groups are dynamically generated and augmented into active set $\mathcal{W}$ for group selection.

In terms of the convergence proof of FGM in \cite{Tan10} and Theorem \ref{theorem:QCQP}, we can obtain the following theorem to illustrate the approximation with an $\epsilon$-optimal solution to the original problem.

\begin{theorem} \label{theorem:eps_optimal}
After Algorithm \ref{algo:two-layer-method} stops in a finite number of steps, the difference between optimal solution $(\textbf{d}^*,\alpha^*)$ of Problem (\ref{prob:svm-mpm-ours_dual_mu})
and the solution $(\textbf{d},\alpha)$ of Algorithm \ref{algo:two-layer-method} is $\mathcal{F}_{\textbf{d}}(\alpha) - \mathcal{F}_{\textbf{d}^*}(\alpha^*) \leq \epsilon$.
\end{theorem}
The detailed proof of Theorem \ref{theorem:eps_optimal} is given in the supplementary material.

\subsection{Finding the Most Violated $\overline{y}'$ and $\textbf{d}$}\label{sec:violate}

Algorithm \ref{algo:group-feature-selection} and  Algorithm\ref{algo:two-layer-method} need to find the most violated $\overline{y}'$ and $\textbf{d}$, respectively. In this subsection, we discuss how to obtain these quantities efficiently.
Algorithm \ref{algo:group-feature-selection} needs to calculate the subgradient of the empirical risk function $R_{emp}^K (\overline{\textbf{w}})$. Since $R_{emp}^K (\overline{\textbf{w}})$ is a pointwise supremum function, the subgradient should be in the convex hull of the gradient of the decomposed functions with the largest objective. Here, we just take one of these subgradients by solving
\begin{eqnarray}
\overline{y}^k = \arg \max_{\overline{y}' \in \overline{\mathcal{Y}} \backslash \overline{y}} \Delta(\overline{y}',\overline{y}) - \sum_{i=1}^n (y_i - y_i') v_i, \label{prob:find-violated-y}
\end{eqnarray}
where $v_i = \sum_{t=1}^T \textbf{w}_t^T (\textbf{x}_i \odot \sqrt{\textbf{d}^t})$. After obtaining $\overline{y}^k$, it is easy to compute $\textbf{p}^k_t = -\frac{1}{n} \sum_{i=1}^n (y_i - y_i^k)(\textbf{x}_i \odot \sqrt{\textbf{d}^t})$ and $q^k = \frac{1}{n} \sum_{i=1}^n \Delta(\overline{y}^k,\overline{y})$.

For finding the most violated $\overline{y}'$, it depends on how to define the loss $\Delta(\overline{y}, \overline{y}')$ in Problem (\ref{prob:find-violated-y}). One of the instances is the Hamming loss which can be decomposed and computed independently, i.e., $\Delta(\overline{y}, \overline{y}') = \sum_{i=1}^n \delta(y_i,y'_i)$, where $\delta$ is an indicator function with $\delta(y_i,y'_i) = 0$ if $y_i = y'_i$, otherwise $1$. However, there are some multivariate performance measures which could not be solved independently. Fortunately, there are a series of structured loss functions, such as Area Under ROC (AUC), Average Precision (AP), ranking and contingency table scores and other measures listed in \cite{Joachims05,SVM_MAP,Teo10}, which can be implemented efficiently in our algorithms. In this paper, we only use several multivariate performance measures based on contingency table as the showcases and their finding $\overline{y}^k$ could be solved in time complexity $O(n^2)$ \cite{Joachims05}.

Given the true labels $\textbf{y}$ and predicted labels $\textbf{y}'$, the contingency tables is defined as follows

\begin{center}
\begin{footnotesize}
\begin{tabular}{|c|c|c|}
\hline
       & y=1 & y=-1 \\
\hline
y'=1   & a   & b \\
\hline
y'=-1  & c   & d \\
\hline
\end{tabular}
\end{footnotesize}
\end{center}

\textbf{$F_1$-score}: The $F_{\beta}$-score is a weighted harmonic average of Precision and Recall. According to the contingency table, we can obtain
$F_{\beta} = \frac{(1 + \beta^2)a}{(1+\beta^2)a + b + \beta^2 c}.$
The most common choice is $\beta=1$. The corresponding balanced $F_1$ measure loss can be written as $\Delta_{F_1}(a,b,c,d) = 100 (1-F_1)$. Then, Algorithm 2 in \cite{Joachims05} can be directly applied.

\textbf{Precision/Recall@k:} In search engine systems, most users scan only the first few links that are presented. In this situation, Prec@k and Rec@k measure the precision and recall of a classifier that predicts exactly $k$ documents, i.e.,
$Prec@k = \frac{a}{a+b} $ and $ Rec@k = \frac{a}{a+c},$
subject to $a+b = k$. The corresponding loss could be defined as $\Delta_{Prec@k} = 100( 1 - Prec@k)$ and $\Delta_{Rec@k} = 100 (1 - Rec@k)$. And the procedure of finding most violated $\textbf{y}$ is similar to F-score, while the only difference is keeping constraint $a+b = k$ and removing $a+b \not = k$.

\textbf{Precision/Recall Break-Even Point (PRBEP):} The Precision/Recall Break-Even Point requires that the precision and its recall are equal. According to above definition, we can see PRBEP only adds a constraint $a+b = a+c$, or $b = c$. The corresponding loss is  defined as $\Delta_{PRBEP} = 100 (1 - PRBEP)$. Finding the most violated $\textbf{y}$ should enforce the constraint $b = c$.

After $t$ iterations in Algorithm \ref{algo:two-layer-method}, we transform $\alpha$ in Problem (\ref{prob:find_most_violated_d}) from the exponential size to a small size $\alpha^t$. Now, finding the most violated $\textbf{d}$  becomes
\begin{eqnarray}
\textbf{d}^t \!\!\!\!\!&=&\!\!\!\!\! \arg \max_{\textbf{d} \in \mathcal{D}} \mathcal{G}_{\textbf{d}}(\alpha^t) \label{prob:find-violated-d-shrink}\\
             \!\!\!\!\!&=&\!\!\!\!\! \arg \max_{\textbf{d} \in \mathcal{D}} \frac{1}{2} \bigg\|\sum_{k=1}^K \alpha_k^t \textbf{p}^k\bigg\|_2^2 - \sum_{k=1}^K \alpha_k^t q^k \nonumber\\
             \!\!\!\!\!&=&\!\!\!\!\! \arg \max_{\textbf{d} \in \mathcal{D}} \frac{1}{2} \bigg\|\frac{1}{n} \sum_{k=1}^K \alpha_k^t\sum_{i=1}^n (y_i - y^k_i) (\textbf{x}_i \odot \sqrt{\textbf{d}}) \bigg\|^2 \nonumber\\
             \!\!\!\!\!&=&\!\!\!\!\! \arg \max_{\textbf{d} \in \mathcal{D}} \frac{1}{2 n^2} \sum_{j=1}^m  c_j^2 d_j \nonumber
\end{eqnarray}
where $c_j = \sum_{k=1}^K \alpha_k^t \sum_{i=1}^n (y_i - y^k_i) \textbf{x}_{i,j} $. With the budget constraint $\sum_{i=1}^m d_i = B$ in $\mathcal{D}$,  (\ref{prob:find-violated-d-shrink}) can be solved by first sorting $c_j^2$'s in the descent order and then setting the first $B$ numbers corresponding to $d_j^t$ to $1$ and the rest to $0$. This takes only $O(m \log m)$ operations.

\section{Relations To Existing Methods} \label{sec:relationship}

In this section, we will discuss the relationships between our proposed method for multivariate loss (\ref{prob:svm-simplemkl-general}) and the state-of-the-art feature selection methods including SKM \cite{Bach04}, NMMKL \cite{Xu09}, $l_1$-SVM \cite{YuanLR2009}, $l_0$-SVM \cite{Weston03} and FGM \cite{Tan10}. It can be easily adapted to the general framework (\ref{prob:svm-simplemkl-framework}).

\subsection{Connections to SKM and $l_1$ SVM}

Let $\mathcal{D}_1 = \{ \mathbf{d} | \sum_{j=1}^m d_j = 1, d_j \geq 0, \forall j=1,\ldots,m\}$ be in the real domain. We observe that $\mathcal{D} = \mathcal{D}_1$ when $B=1$. According to \cite{Rakotomamonjy08}, we transform Problem (\ref{prob:svm-simplemkl-general}) in the special case of $B=1$ to the following equivalent optimization problem,
\begin{eqnarray}
\min_{\textbf{w},\xi \geq 0} \!\!\!\!\!&&\!\!\!\!\! \frac{1}{2} \Bigg(\sum_{j=1}^m |w_j| \Bigg)^2 + C \xi \label{prob:svm-skm}\\
\textrm{s.t.} \!\!\!\!\!&&\!\!\!\!\! \forall \overline{y}' \in \overline{\mathcal{Y}} \backslash \overline{y} : {\textbf{w}}^T \frac{1}{n} \sum_{i=1}^n (y_i - y'_i) \textbf{x}_i \geq \Delta(\overline{y},\overline{y}' ) - \xi. \nonumber
\end{eqnarray}
SKM \cite{Bach04} attempts to obtain the sparsity of $\mathbf{w}$ by penalizing the square of a weighted block $l_1$-norm $(\sum_{j=1}^k \gamma_j ||\textbf{w}_j||_2)^2$ where $k$ is the number of groups and $\mathbf{w}_j$ is the weight vector for the features in the $j$th group. The regularizer $(\sum_{j=1}^m |w_j|)^2$ used in (\ref{prob:svm-skm}) is the square of the $l_1$ norm $(||\mathbf{w}||_1)^2$, which is a special case of SKM when $k=m$ and $\gamma_j=1$, i.e., each group contains only one feature. Minimizing the square of the $l_1$-norm is very similar to $l_1$-norm SVM \cite{YuanLR2009} by setting $\Omega(\mathbf{w}) = ||\mathbf{w}||_1$ with the non-negative (convex) loss function.

Regardless of $l_1$-norm or the square of $l_1$-norm, the parameter $C$ is too sensitive to be tuned properly since it controls both margin loss and the sparsity of $\mathbf{w}$. However, our method alleviates this problem by two parameters $C$ and $B$ which control margin loss and sparsity of $\mathbf{w}$, respectively. This separation makes parameter tuning of our method easier than those of SKM and $l_1$ SVM.

\subsection{Connection to NMMKL}

Instead of directly solving Problem (\ref{prob:svm-skm}), we formulate a more general problem (\ref{prob:svm-simplemkl-general}) by introducing an additional budget parameter $B$, which directly controls the sparsity of $\mathbf{w}$. The advantage is to make parameter tuning easily done since $C$ is not sensitive to the sparsity of $\mathbf{w}$. This strategy is also used in NMMKL \cite{Xu09}, but one feature corresponding to one base kernel makes NMMKL intractable for high-dimensional problems. The multivariate loss is even hard to be optimized by NMMKL since there are exponential dual variables in the dual form of NMMKL from the exponential number of constraints. However, our method can readily optimize multivariate loss on high-dimensional data.

\subsection{Connection to FGM}

According to the work \cite{Zien2007}, we can reformulate Problem (\ref{prob:svm-skm}) as an equivalent optimization problem
\begin{eqnarray}
\min_{\mathbf{d} \in \mathcal{D}_1}\min_{\textbf{w},\xi \geq 0} &&\!\!\!\!\!\!\! \frac{1}{2} \sum_{j=1}^m d_j |w_j|^2 + C \xi  \label{prob:svm-mcmkl}\\
\textrm{s.t.} \forall \overline{y}' \in \overline{\mathcal{Y}} \backslash \overline{y}\!\!\!\!\!&&\!\!\!\!\!  : \frac{1}{n} \sum_{j=1}^m d_j w_j \sum_{i=1}^n (y_i - y'_i) \textbf{x}_{j,i} \geq \Delta(\overline{y},\overline{y}' ) - \xi. \nonumber
\end{eqnarray}
After the substitutions of $v_j = \sqrt{d}_j w_j, \forall j=1,\ldots,m$ and the general case of $\mathcal{D}$, we can obtain the following problem
\begin{eqnarray}\label{prob:svm-mpm-ours}
\min_{\textbf{d} \in \mathcal{D}} \min_{\textbf{v},\xi \geq 0} &&\!\!\!\!\! \frac{1}{2} \|\textbf{v}\|^2_2 + C \xi \\
\textrm{s.t.} \forall \overline{y}' \in \overline{\mathcal{Y}} \backslash \overline{y}\!\!\!\!\! &&\!\!\!\!\! : \textbf{v}^T \frac{1}{n} \sum_{i=1}^n (y_i - y'_i)  (\textbf{x}_i \odot \sqrt{\textbf{d}}) \geq \widetilde{\Delta}(\overline{y},\overline{y}' ) - \xi,\nonumber
\end{eqnarray}
where $\mathbf{v}=[v_1,\ldots,v_m]^T$. After deriving Lagrangian dual problem of (\ref{prob:svm-mpm-ours}), we observe that it is same as Problem (\ref{prob:our-dual}). Problem (\ref{prob:find-violated-d-shrink}) always finds the most violated $\mathbf{d}$ in the integer domain $\{0,1\}^m$, so the solutions of the following problem solved by the proposed two-layer cutting plane algorithm is the same as the solutions of Problem (\ref{prob:our-dual})
\begin{eqnarray}\label{prob:svm-mpm-ours-d}
\min_{\textbf{d} \in \mathcal{D}_2} \min_{\textbf{v},\xi \geq 0} &&\!\!\!\!\! \frac{1}{2} \|\textbf{v}\|^2_2 + C \xi \\
\textrm{s.t.} \forall \overline{y}' \in \overline{\mathcal{Y}} \backslash \overline{y}\!\!\!\!\! &&\!\!\!\!\! : \textbf{v}^T \frac{1}{n} \sum_{i=1}^n (y_i - y'_i)  (\textbf{x}_i \odot \textbf{d}) \geq \widetilde{\Delta}(\overline{y},\overline{y}' ) - \xi,\nonumber
\end{eqnarray}
where the integer domain $\mathcal{D}_2 = \{ \mathbf{d} | \sum_{j=1}^m d_j \leq B, \mathbf{d} \in \{0,1\}^m\}$. This formula can be equally derived as the extension of FGM for multivariate performance measures by defining the new hypotheses
\begin{equation}
\widetilde{h}_{\textbf{v}}(\textbf{x}) = \arg \max_{\overline{y}' \in \overline{\mathcal{Y}}}  \sum_{i=1}^n y'_i ( \textbf{v}  \odot \textbf{d})^T \textbf{x}_i, \label{eq:ldf-ours}
\end{equation}
where $\widetilde{h}_{\textbf{v}} : \overline{\mathcal{X}} \rightarrow \overline{\mathcal{Y}}$ and $\mathbf{d} \in \mathcal{D}_2$. It is not trivial to perform the extension of FGM to optimize multivariate loss because original FGM method \cite{Tan10} cannot directly apply to solve the exponential number of constraints. And our domain of $\mathbf{d}$ is in real domain $\mathcal{D}$ which is more general than the integer domain $\mathcal{D}_2$ used in FGM and the proposed extension (\ref{prob:svm-mpm-ours-d}), even though the final solutions of (\ref{prob:svm-simplemkl-general}) and (\ref{prob:svm-mpm-ours-d}) are the same.

\subsection{Connection to $l_0$ SVM}

The following Lemma indicates that the proposed formula can be interpreted by $l_0$-norm constraint.
\begin{lemma}\label{lemma:equ}
(\ref{prob:svm-mpm-ours-d}) is equivalent to the following problem
\begin{eqnarray}\label{eq:DCSVM}
\!\!\!\!&\!\!\min_{\tilde{\textbf{w}},\xi \geq 0} \!\!& \!\!\frac{1}{2} \|\tilde{\textbf{w}}\|^2_2 + C \xi \\
\!\!\!\!\!\!\!\!\!&\!\!\!\!\textrm{s.t.} \!\!\!\!&\!\!\!\forall \overline{y}' \in \overline{\mathcal{Y}} \backslash \overline{y} : \tilde{\textbf{w}}^T \frac{1}{n} \sum_{i=1}^n (y_i - y'_i)  \textbf{x}_i  \geq \widetilde{\Delta}(\overline{y},\overline{y}' ) - \xi,\nonumber\\
\!\!\!\!\!\!\!\!\!&\!\!\!\!\!&\|\tilde{\textbf{w}}\|_0 \leq B.
\nonumber
\end{eqnarray}
\end{lemma}
\begin{proof}
Note,  at the optimality of (\ref{prob:svm-mpm-ours}),
WLOG,  suppose $d_j=0$, the corresponding $v_j$ must be 0.
Thus, $\|{\textbf{v}}\|_0 \leq \|\textbf{d}\|_0$. Let $\tilde{\textbf{w}}=\textbf{v}\odot \textbf{d}$,  we have
$\|\tilde{\textbf{w}}\|_0 = \|{\textbf{v}}\odot \textbf{d}\|_0\leq \min\{\|{\textbf{v}}\|_0,
\|\textbf{d}\|_0\}\leq\|\textbf{d}\|_0=\sum_{j=1}^m d_j\leq B.$ Moreover,
$\|\tilde{\textbf{w}}\|_2^2=\|{\textbf{v}}\odot \textbf{d}\|_2^2=\|{\textbf{v}}\|_2^2$ at the optimality.
Therefore,  the optimal solution of (\ref{prob:svm-mpm-ours}) is a feasible
solution of (\ref{eq:DCSVM}). On the other hand,  for the optimal
$\tilde{\textbf{w}}$ in (\ref{eq:DCSVM}),  let ${\textbf{v}}=\tilde{\textbf{w}}$  and $d_i=\delta(\tilde{w_i})$ where
$\delta(t)=1$ if $t\neq 0$; otherwise,  0. So,  the optimal solution
of (\ref{eq:DCSVM}) is a feasible solution of (\ref{prob:svm-mpm-ours}).
\end{proof}

This gives another interpretation of parameter $B$ from the perspective of $l_0$-norm. Since $l_0$-norm $||\widetilde{\mathbf{w}}||_0$ represents the number of non-zero entries of $\widetilde{\mathbf{w}}$, so $B$ in our method can be considered as the parameter which directly controls the sparsity of $\mathbf{w}$.

\section{Multiple Instance Learning for Multivariate Performance Measures} \label{sec:mil-multi-variate-measure}

We have already illustrated the proposed framework by optimizing multivariate performance measures for feature selection in Section \ref{sec:feature-selection-for-multivariate}. In this section, we extend this approach to solve multiple instance learning problems which have been employed to solve a variety of learning problems, e.g., drug activity prediction \cite{Dietterich1997}, image retrieval \cite{Zhang2002}, natural scene classification \cite{Maron98} and text categorization \cite{Andrews2003}, but it is seldom optimized for multivariate performance measures in the literature. However, it is crucial to optimize the task specific performance measures, e.g., $F$ score is widely considered as the most important evaluation criterion for a learning method in image retrieval.

Multi-instance learning was formally introduced in the context of drug activity prediction \cite{Dietterich1997}. In this learning scenario, a bag is represented by a set of instances where each instance is represented by a feature vector. The classification label is only assigned to each bag instead of the instances in this bag. We name a bag as a positive bag if there is at least one positive instance in this bag, otherwise it is called negative bag. The learning problem is to decide whether the given unlabeled bag is positive or not. By defining a similarity measure between a bag and an instance, Multiple-Instance Learning via Embedded instance Selection (MILES) \cite{Chen06} successfully transforms multiple instance learning into a feature selection problem by embedding bags into an instance-based feature space and selecting the most important features.

Before discussing the transformation in MILES, we first give the notations of multiple instance learning problem. Following the notations in \cite{Chen06}, we denote $i$th positive bags as $\mathbf{B}^+_i = \{ \mathbf{x}_{i,j}^+ \}_{j=1}^{n_i^+}$ which consists of $n_i^+$ instances $\mathbf{x}_{i,j}^+, j=1,\ldots,n_i^+$. Similarly, the $i$th negative bags is denoted as $\mathbf{B}_i^- = \{\mathbf{x}_{i,j}^-\}_{j=1}^{n_i^-}$. All instances belongs to the same feature space $\mathcal{X}$. The number of positive bags and negative bags are $\ell^+$ and $\ell^-$, respectively. The instances in all bags are rearranged as $\{\mathbf{x}^1,\ldots,\mathbf{x}^n\}$ where $n=\sum_{i=1}^{\ell^+} n_i^+ + \sum_{i=1}^{\ell^-} n_i^-$.

By considering each instance in the training bags as a candidate for target concepts, the embedded feature space is represented as
\begin{equation}
\widehat{\mathbf{x}}_i = [s(\mathbf{x}^1,\mathbf{B}_i),\ldots,s(\mathbf{x}^n,\mathbf{B}_i)]^T \in R^n,
\end{equation}
where the similarity measure between the bag $\mathbf{B}_i$ and the instance $\mathbf{x}^k$ is defined as the most-likely-cause estimator
\begin{equation}
s(\mathbf{x}^k,\mathbf{B}_i) = \max_{j} \exp \left(  - \frac{||\mathbf{x}_{i,j} - \mathbf{x}^k||^2}{2 \sigma^2} \right).
\end{equation}
It follows the intuition that the similarity between a concept and a bag is determined by the concept and the closest instance in this bag. The corresponding labels are constructed as follows: $\widehat{y}_i = 1$ if $\mathbf{B}_i$ is a positive bag, otherwise $\widehat{y}_i = -1$. For a given $\ell^+$ positive bags and $\ell^-$ negative bags, we form a new classification representation of the multiple instance learning problem as $\{ \widehat{\mathbf{x}}_i,\widehat{y}_i\}_{i=1}^{\ell^++\ell^-}$. For each instance $\mathbf{x}^k$, the new feature representation corresponds to the values of the $k$th feature variable $s(\mathbf{x}^k,\cdot)$ is
\begin{displaymath}
[ s(\mathbf{x}^k, \mathbf{B}_1^+),\ldots,s(\mathbf{x}^k, \mathbf{B}_{\ell^+}^+),s(\mathbf{x}^k, \mathbf{B}_1^-),\ldots,s(\mathbf{x}^k, \mathbf{B}_{\ell^-}^-) ]
\end{displaymath}
where the feature induced by $\mathbf{x}^k$ provides the useful information for separating the positive and negative bags. The linear discriminant function
\begin{equation}
\widehat{y} = \textrm{sign} (\langle \mathbf{w}, \widehat{\textbf{x}} \rangle + b)
\end{equation}
where $\mathbf{w}$ and $b$ are the model parameters. The embedding induces a possible high-dimensional space when the number of instances in the training set is large. Since some instances may not be responsible for the label of the bags or might be similar to each other, many features are redundant or irrelevant, so MILES employs $L_1$-SVM to select a subset of mapped features that is most relevant to the classification problem. However, $L_1$-SVM cannot fulfill to obtain a high performance over the task-specific measures because it only focuses on optimizing zero-one loss function. Our proposed Algorithm~\ref{algo:two-layer-method} is a natural alternative feature selection method for multi-variate performance measures. The proposed algorithm for multiple instance learning to optimize multivariate measures is shown in Algorithm \ref{algo:bag-classifier}.

\begin{algorithm}
   \caption{Learning a bag classifier}
   \label{algo:bag-classifier}
\begin{algorithmic}[1]
    \STATE Input: positive bags $\{\mathbf{B}_i^+\}_{i=1}^{\ell^+}$, negative bags $\{\mathbf{B}_i^-\}_{i=1}^{\ell^-}$, $C$, and $\epsilon$
    \STATE Construct the embedding representation of training data $$\{ (\widehat{\mathbf{x}}_i, \widehat{y}_i)\}, \forall i=1,\ldots,\ell^++\ell^-$$
    \STATE $\overline{\mathbf{x}} = [\widehat{\mathbf{x}}_1,\ldots,\widehat{\mathbf{x}}_{\ell^++\ell^-}]$ and $\overline{\mathbf{y}} = [\widehat{y}_1,\ldots,\widehat{y}_{\ell^++\ell^-}]$
    \STATE call Algorithm \ref{algo:two-layer-method} with arguments ($\overline{\mathbf{x}}$,$\overline{\mathbf{y}}$,$C$,$\epsilon$)
    \STATE Output: parameters $\mathbf{w}$
\end{algorithmic}
\end{algorithm}

According to Algorithm \ref{algo:bag-classifier}, we do not need the model parameter $b$ since the structural SVM is irrelevant to the relative offset $b$, i.e., $\widehat{y} = \arg \max_{y \in \{-1,+1\}} y \langle \widehat{\mathbf{w}}, \widehat{\mathbf{x}} \rangle$.

\section{Experiments} \label{sec:experimetns}

\begin{table}
\caption{Datasets used in our experiments} \label{tab:datasets}
\begin{center}
\begin{footnotesize}
\begin{tabular}{l|crrr}
\hline
Dataset  & \#classes & \#features & \#train & \#test \\
         &           &              & points & points \\
\hline
\!News20.binary\!\!\!    & 2   & 1,355,191 &11,997  & 7,999  \\
\!URL1             & 2   &3,231,961  &20,000  &20,000 \\
\hline
\!Image            & 5    &10,800   & 1,200 & 800 \\
\hline
\!Sector           & 105 & 55,197    &6,412   & 3,207 \\
\!News20           & 20  & 62,061    &15,935  & 3,993 \\
\hline
\end{tabular}
\end{footnotesize}\noindent
\end{center}
\end{table}

In this Section, we conduct extensive experiments to evaluate the performance of
our proposed method and state-of-the-art feature selection methods:
1) SVM-RFE~\cite{Guyou02}; 2) $l_1$-SVM; 3)  FGM~\cite{Tan10}; 4) $l_1$-bmrm-F$_1$\footnote{http://users.cecs.anu.edu.au/\~{}chteo/BMRM.html}, which is $l_1$ regularized SVM for optimizing F$_1$ score by bundle method~\cite{Teo10}.
SVM-RFE and FGM use Liblinear software \footnote{http://www.csie.ntu.edu.tw/\~{}cjlin/liblinear/} as the QP solver for their SVM subproblems. For $l_1$-SVM, we also use Liblinear software, which implements the state-of-the-art $l_1$-SVM algorithm~\cite{YuanLR2009}.
In addition to the comparison for $0$-$1$ loss, we also perform experiments on image data for F1 measure. Furthermore, several specific measures on the contingency table are investigated on Text datasets by comparing with $\textrm{SVM}^{perf}$ \cite{Joachims05}. All the datasets shown in Table~\ref{tab:datasets} are of high dimensions.

For convenience, we name our proposed two-layer cutting plane algorithm $\textrm{FS}_{multi}^{\Delta}$, where $\Delta$ represents different type of multivariate performance measures. We implemented Algorithm \ref{algo:two-layer-method} in MATLAB for all the multivariate performance measures listed above, using Mosek as the QCQP solver for Problem (\ref{prob:small-QCQP}) which yields a worse-case complexity of $O(KT^2)$. Removing inactive constraints from the working set ~\cite{Joachims09} in the inner layer is employed for speedup the QCQP problem. Since the values of both $K$ and $T$ are much smaller than the number of  examples $n$ and its dimensionality $m$, the QCQP  is very efficient as well as more accurate for large-scale and high-dimensional datasets. Furthermore, the codes simultaneously solve the primal and its dual form. So the
optimal $\mu$ and $\alpha$ can be obtained  after solving Problem (\ref{prob:small-QCQP}).

For a test pattern $\textbf{x}$, the discriminant function can be obtained by $ f(\textbf{x}) = \langle \textbf{w} \odot \widetilde{\textbf{d}}, \textbf{x} \rangle $ where $\textbf{w} = \sum_{i=1}^n \beta_i \textbf{x}_i$, $\beta_i = \frac{1}{n} \sum_{k=1}^K \alpha_k (y_i - y_i^k)$, and $\widetilde{\textbf{d}} = \sum_{t=1}^T \mu_t \sqrt{\textbf{d}^t}$. This leads to the faster prediction since only a few of the selected features are involved. After computing $\textbf{p}^k$, the matrices of Problem (\ref{prob:small-QCQP}) can be incrementally updated, so it can be done totally in $O(T K^2)$.

\begin{figure}
\centering
\begin{tabular}{cc}
\includegraphics[width=0.5\textwidth]{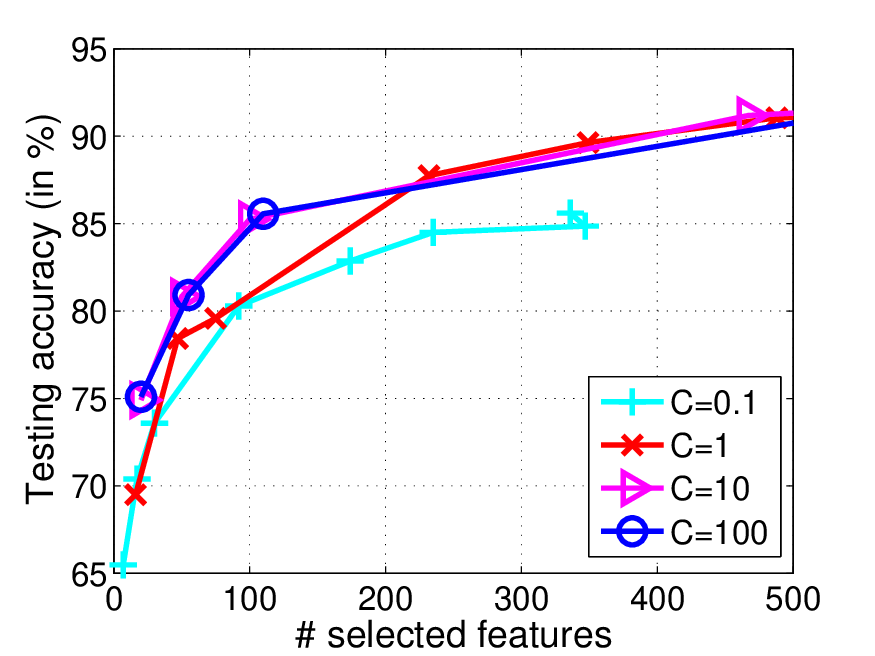}\!\!\!\!\!\! & \!\!\!\!\!\!\includegraphics[width=0.5\textwidth]{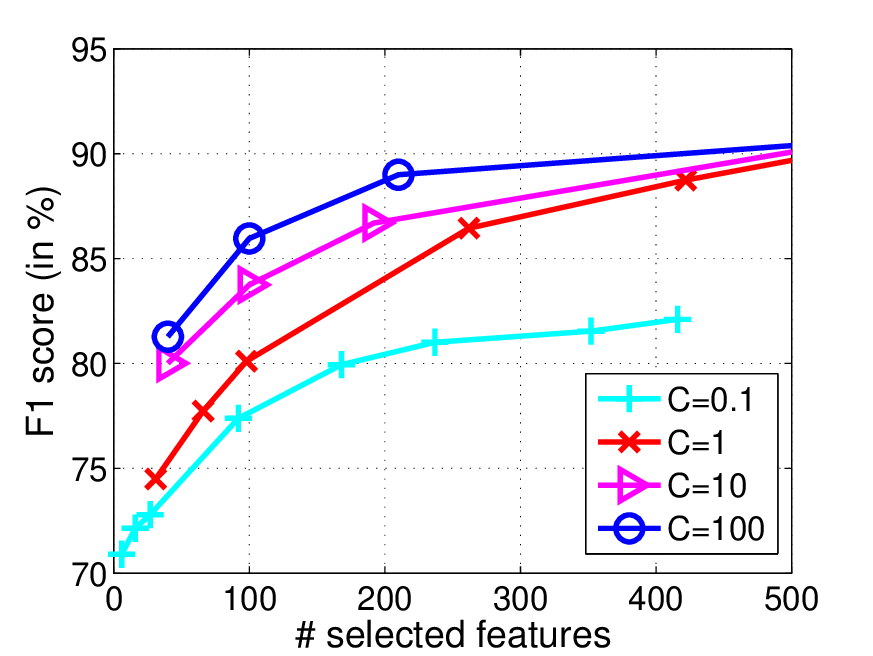}\\
(a) \!\!\!\!\!\! & \!\!\!\!\!\! (b)\\
\includegraphics[width=0.5\textwidth]{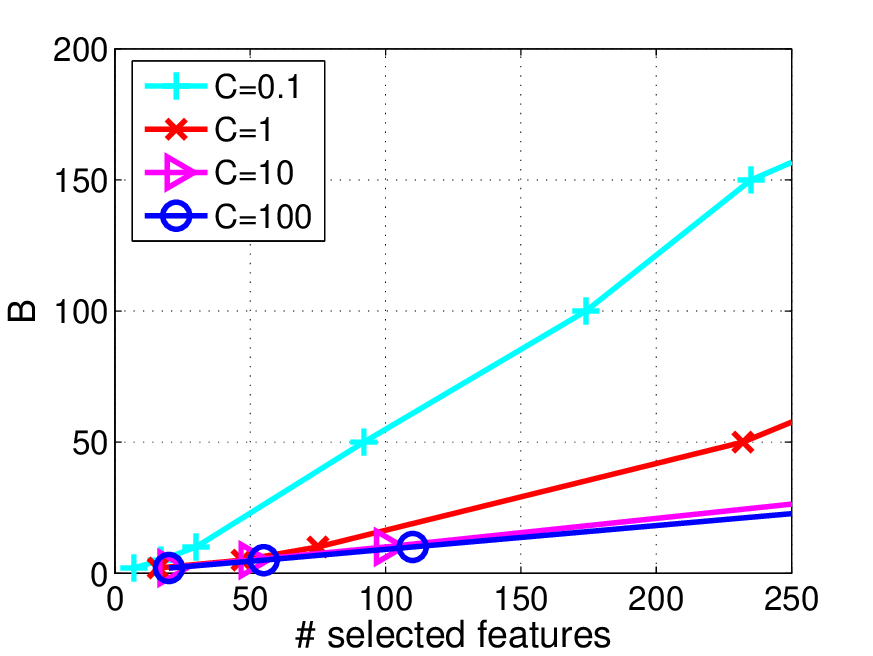}\!\!\!\!\!\! & \!\!\!\!\!\!\includegraphics[width=0.5\textwidth]{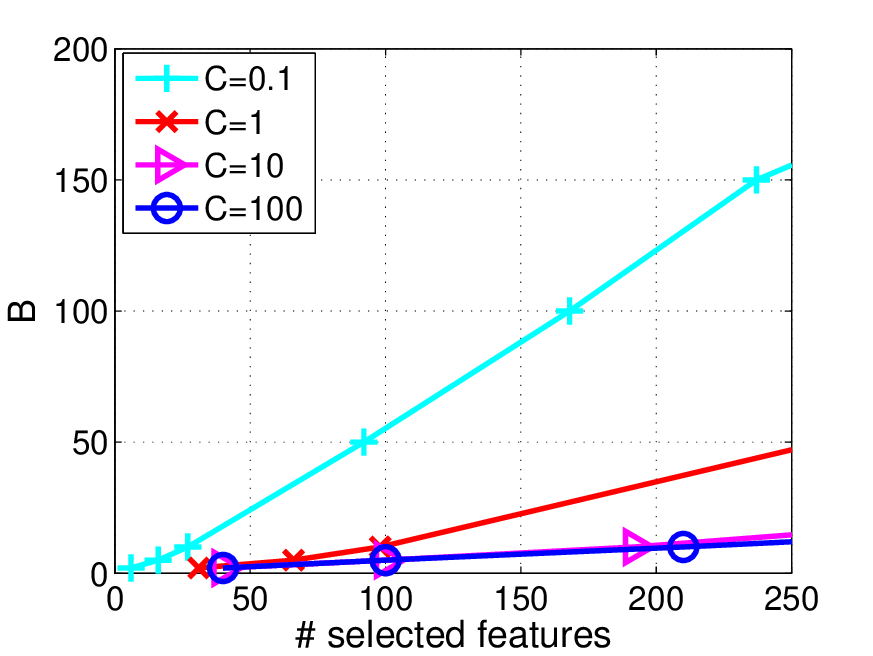}\\
(c) \!\!\!\!\!\! & \!\!\!\!\!\! (d)
\end{tabular}
\caption{ First row (a-b): Testing Accuracy and $F_1$ scores as well as the number of selected features of the proposed method $\textrm{FS}_{multi}^{\Delta}$ on \emph{News20.binary} dataset by varying $C$ and $B$. Second row (c-d): The corresponding relationship among parameters $B$, $C$, and the number of selected features.}\label{fig:change-C-B}
\end{figure}

\subsection{Parameter Sensitivity Analysis} \label{sec:parameter-sensitive-analysis}

Before comparing $\textrm{FS}^{\Delta}_{multi}$ with other methods, we first conduct empirical studies for the parameter sensitivity analysis on \emph{News20.binary}. The goal is to examine the relationships among parameters $C$ and $B$, performance measures and the number of selected features with the range of $C$ in $[0.1, 1, 10, 100] \times n$ and $B$ in $[2, 5, 10, 50, 100, 150, 200, 250]$.

Figure \ref{fig:change-C-B}(a-b) show the testing accuracy and F1 scores as well as the number of selected features by varying $C$ and $B$. We observe that the results are very sensitive to $C$ when $B$ is very small. This indicates that the $l_1$ model, which is equivalent to the proposed method in the case of $B=1$, is vulnerable to the choice of $C$. On the other hand, the results are rather insensitive to $C$ when $B$ is large. Hence, the proposed method is less sensitive to $C$ than $l_1$ model. We also observe that the proposed method prefers a large $C$ value for better performances.
Figure \ref{fig:change-C-B}(c-d) demonstrate the corresponding relationships  among parameters $B$, $C$ and the number of selected features of Figure \ref{fig:change-C-B}(a-b). We observe that  $B$ and the number of selected features always exhibits a linear trend with a constant slope. Moreover, the slope remains the same when $C\geq 10$, but a small $C$ will increase the slope. This means that, compared with $B$, parameter $C$ has less influence on the sparsity of $\mathbf{w}$, and
the learned feature selection model  becomes stabilized when $C\geq 10$.
These empirical results are consistent to the discussions of parameter $B$ in Section \ref{sec:generalized-spase-regularizer}.

 Since large $C$ needs more iterations to converge according to Theorem \ref{theorem:QCQP}, the compromise is to set $C$ not too large and let $B$ dominate the selection of features.
According to these observations, we can safely fix $C$ and study the results by varying $B$ to compare with other methods in the following experiments.

\subsection{Time Complexity Analysis} \label{sec:time}

\begin{figure}
\centering
\begin{tabular}{cc}
\!\!\!\includegraphics[width=0.45\textwidth]{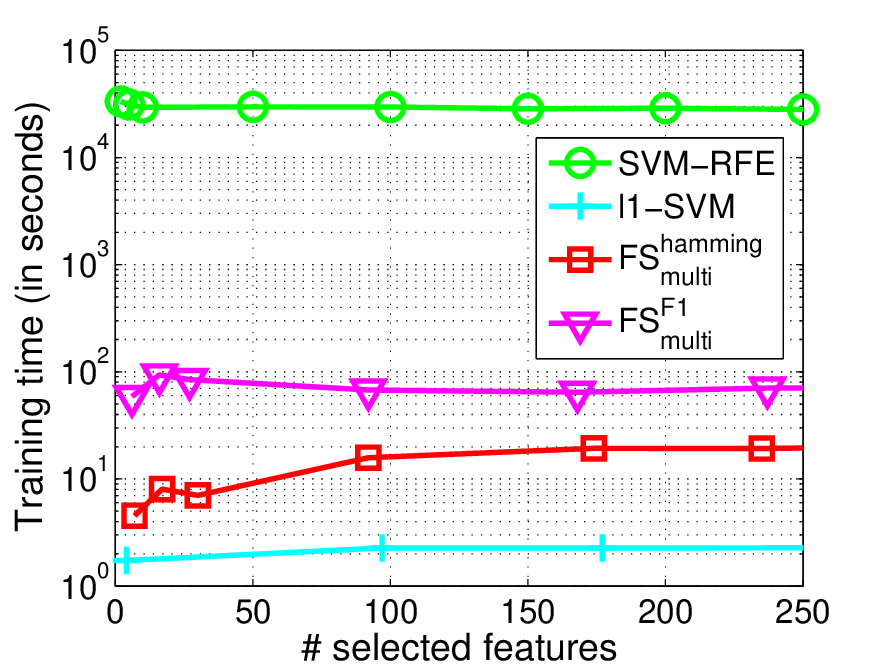}\!\!\!\!\!\!&\!\!\!\!\!\!
\includegraphics[width=0.45\textwidth]{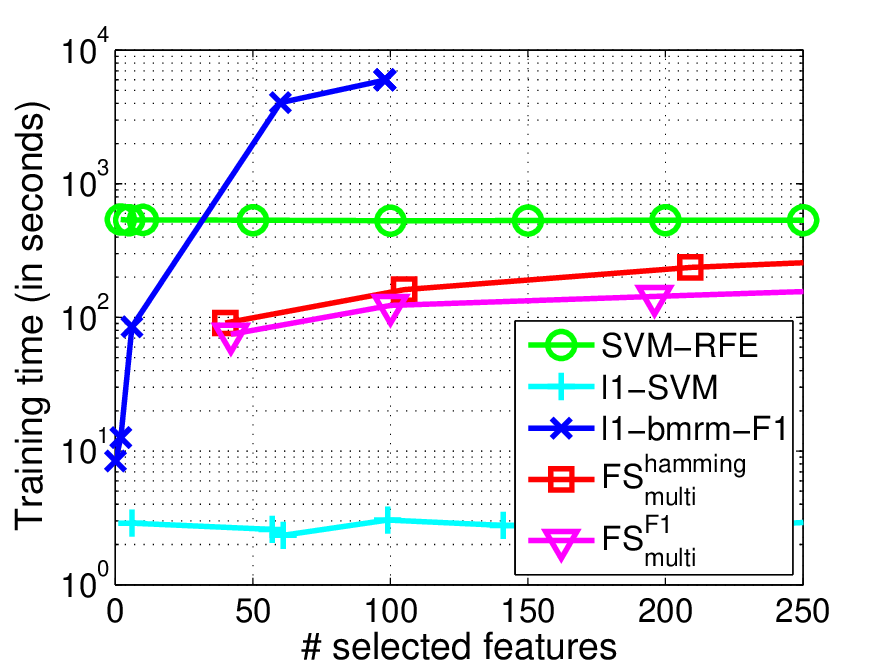}\\
(a) News20.binary\!\!\!\!\!\!&\!\!\!\!\!\!(b) Image (Desert)\\
\end{tabular}
\caption{ Training time on different datasets }\label{fig:training-time-hamming}
\end{figure}

We empirically study the time complexity of $\textrm{FS}_{multi}^{F_1}$ by comparing with other methods. Two datasets \emph{News20.binary} and \emph{Image (Desert)} are used for illustration. The detailed setting are shown in Section \ref{sec:fs-accuracy} and Section \ref{sec:fs-image}, respectively. Figure \ref{fig:training-time-hamming} gives the training time over five different methods. On \emph{News20.binary} dataset, we cannot report the training time for
$l_1$-bmrm-F$_1$
since $l_1$-bmrm-F$_1$ cannot terminate after more than two days with the maximum iteration $1000$ and parameter $\lambda \in [10^{-7},10^2]$ due to the extremely high dimensionality.
We observe that the proposed methods are slower than $l_1$-SVM, but much faster than SVM-RFE and $l_1$-bmrm-F$_1$. In addition, on \emph{Image} dataset, when  the termination condition with the relative difference between the objective and its convex linear lower bound lower than $0.1$ is set, $l_1$-bmrm-F$_1$ also cannot converge after the maximum iteration, which is consistent with the discussion in Appendix C of~\cite{Teo10} that bundle method with $l_1$ regularizer cannot guarantee the convergence.
This leads to the similar number of selected features (e.g., $98$ in Figure \ref{fig:training-time-hamming}(b))  even though $\lambda$ is decreasing gradually.

These observations implies that our proposed two-layer cutting plane method needs less time for training with guaranteed convergence than bundle method. Moreover, our method can work on large scale and high dimensional data for optimizing user-specified measure, but bundle method cannot. As aforementioned, $l_1$-bmrm-F$_1$ is much slower on the high dimensional datasets in our experiments, so we can only report its results  in Section \ref{sec:fs-image}.

\subsection{Feature Selection for Accuracy} \label{sec:fs-accuracy}

Since \cite{Joachims05} has proven that $\textrm{SVM}_{multi}^{\Delta}$ with Hamming loss, namely $\Delta_{Err}(\overline{y},\overline{y}') = 2 (b + c)$, is the same as SVM.
In this subsection, we evaluate the accuracy performances of $\textrm{FS}_{multi}^{\Delta}$
for Hamming loss function, namely $\textrm{FS}_{multi}^{hamming}$ as well as other state-of-the-art feature selection methods. We compare these methods on two binary datasets,
\emph{News20.binary}~\footnote{http://www.csie.ntu.edu.tw/\~{}cjlin/libsvmtools/datasets} and \emph{URL1} in Table \ref{tab:datasets}. Both datasets are used in \cite{Tan10},
and they are already split into training and testing sets.

We test FGM and SVM-RFE in the grid $C_{FGM} = [0.001, 0.01, 0.1, 1, 5, 10]$ and choose  $C_{FGM}=5$ which gives good performance for both FGM and SVM-RFE. This is the same as \cite{Tan10}. For $\textrm{FS}_{multi}^{hamming}$, we do the experiments by fixing $C_{FGM_{multi}} $ as $ 0.1 \times n$ for \emph{URL1} and $1.0 \times n$ for \emph{New20.binary}. The setting for budget parameter $B = [2, 5, 10, 50, 100, 150, 200, 250]$ for \emph{News20.binary}, and $B=[2, 5, 10, 20, 30, 40, 50, 60]$ for \emph{URL1}. The elimination scheme of features for SVM-RFE method can be referred to \cite{Tan10}. For $l_1$-SVM, we report the results of different $C$ values so as to obtain different number of selected features.

\begin{figure}
\centering
\begin{tabular}{cc}
\!\!\!\includegraphics[width=0.45\textwidth]{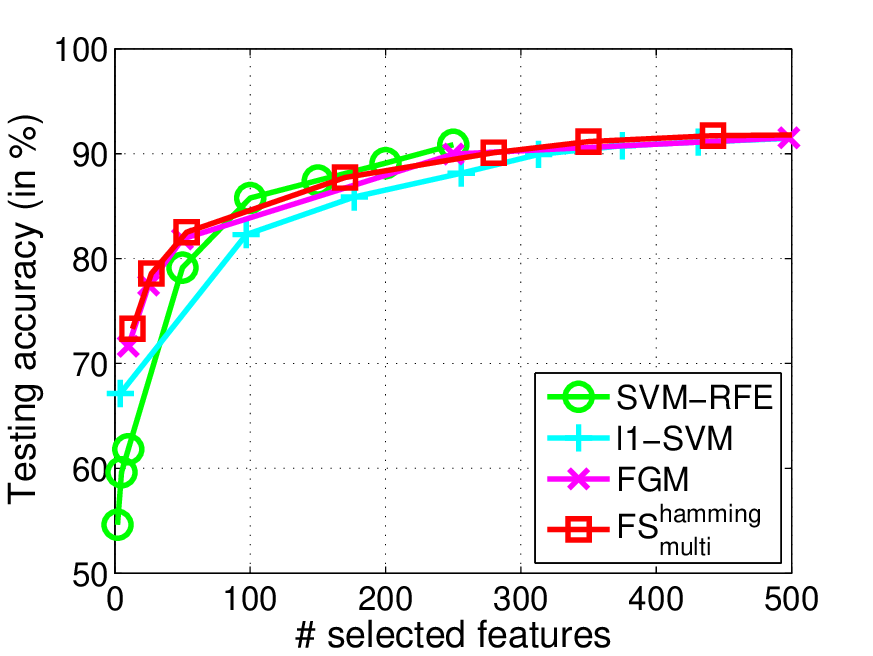}\!\!\!\!\!\!&\!\!\!\!\!\!
\includegraphics[width=0.45\textwidth]{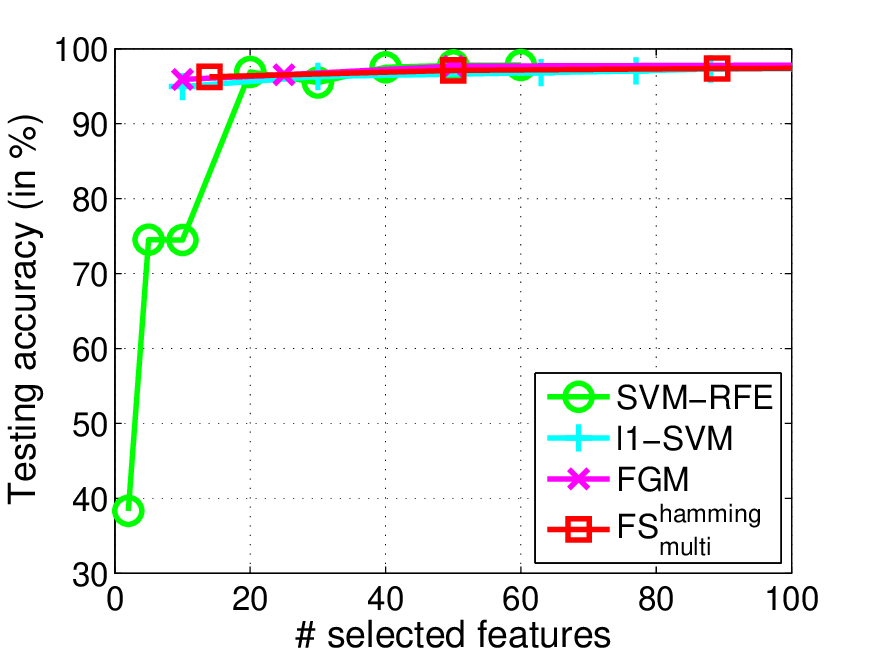}\\
(a) News20.binary\!\!\!\!\!\!&\!\!\!\!\!\!(b) URL1\\
\end{tabular}
\caption{ Testing accuracy on different datasets }\label{fig:testing-accuracy-hamming}
\end{figure}

\begin{figure}
\centering
\begin{tabular}{cc}
\!\!\!\includegraphics[width=0.45\textwidth]{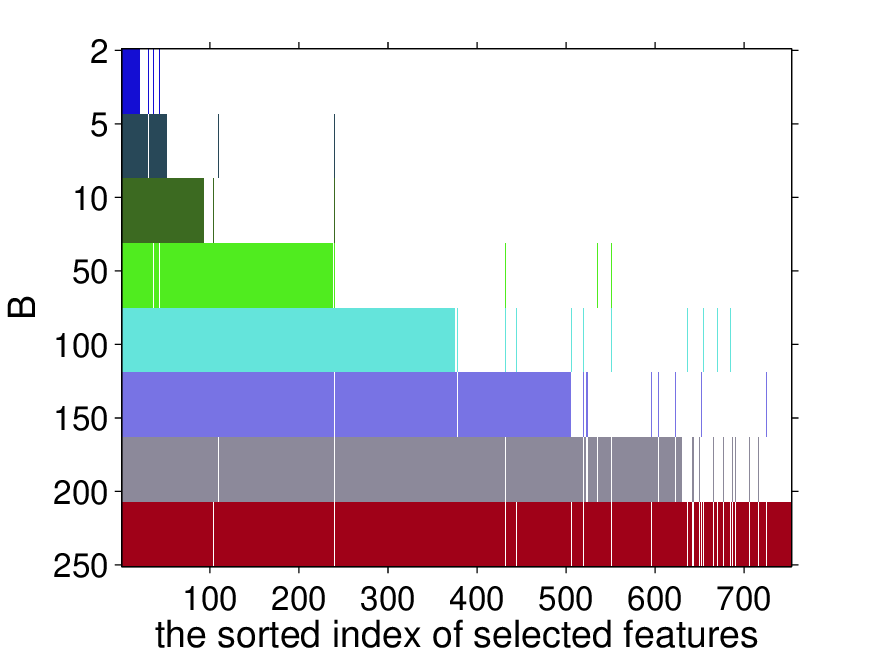}
\!\!\!\!\!\!\!&\!\!\!\!\!\!\!
\includegraphics[width=0.45\textwidth]{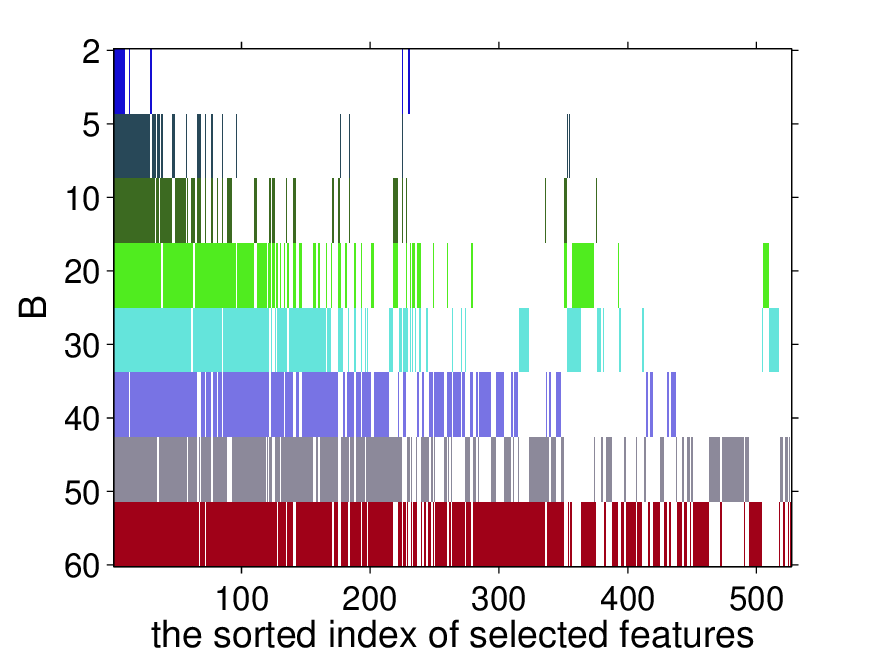}\\
(a) News20.binary\!\!\!\!\!\!&\!\!\!\!\!\!(b) URL1\\
\end{tabular}
\caption{ The sparsity of features of $\textrm{FS}_{multi}^{hamming}$ with varying $B$   on different datasets. Each row bar with different color represents the different subset of features selected under current $B$, where the white region means the features are not selected. }\label{fig:sparse-hamming}
\end{figure}

Figure \ref{fig:testing-accuracy-hamming}  reports testing accuracy on different datasets. The testing accuracy is comparable among different methods, but both $\textrm{FS}_{multi}^{hamming}$ and FGM can obtain better prediction performances than SVM-RFE in a small number (less than 20) of selected features  on both \emph{News20.binary} and \emph{URL1}. These results show that the proposed method with Hamming loss can work well on feature selection tasks especially when choosing only a few features.
$\textrm{FS}_{multi}^{hamming}$ also performs better than  $l_1$-SVM  on \emph{News20.binary} in most range of selected features. This is possibly because $l_1$ models are more sensitive to noisy or redundant features on \emph{News20.binary} dataset.

Figure \ref{fig:sparse-hamming} shows that our method with  the small $B$ will select smaller number of features than the large $B$. We also observed that most of features selected by the small $B$ also appeared in the subset of features using the large $B$. This phenomenon can be obviously observed on \emph{News20.binary}.  This leads to the conclusion that $\textrm{FS}_{multi}^{hamming}$ can select the important features in the given datasets due to the insensitivity of parameter $B$. However, we notice that not all the features in the selected subset of features with smaller $B$ fall into that of subset of features with the large $B$, so our method is non-monotonic feature selection. This argument is consistent with the test accuracy in Figure \ref{fig:testing-accuracy-hamming}. \emph{News20.binary} seems to be monotonic datasets from Figure \ref{fig:sparse-hamming}, since $\textrm{FS}_{multi}^{hamming}$, FGM and SVM-RFE demonstrate similar performance. However, \emph{URL1} is more likely to be non-monotonic, as our method and FGM can do better than SVM-RFE. All the facts imply that the proposed method is comparable with FGM and SVM-RFE. And it also demonstrates the non-monotonic property for feature selection.

\subsection{Feature Selection for Image Retrieval} \label{sec:fs-image}

\begin{figure}
\centering
\begin{tabular}{cc}
\includegraphics[width=0.5\textwidth]{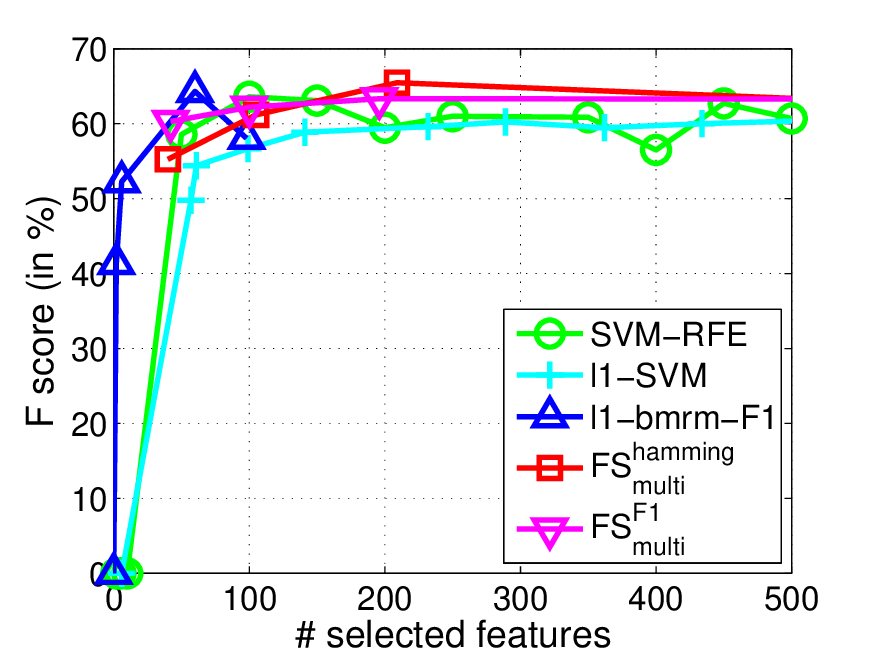}\!\!\!\!\!\! & \!\!\!\!\!\!\includegraphics[width=0.5\textwidth]{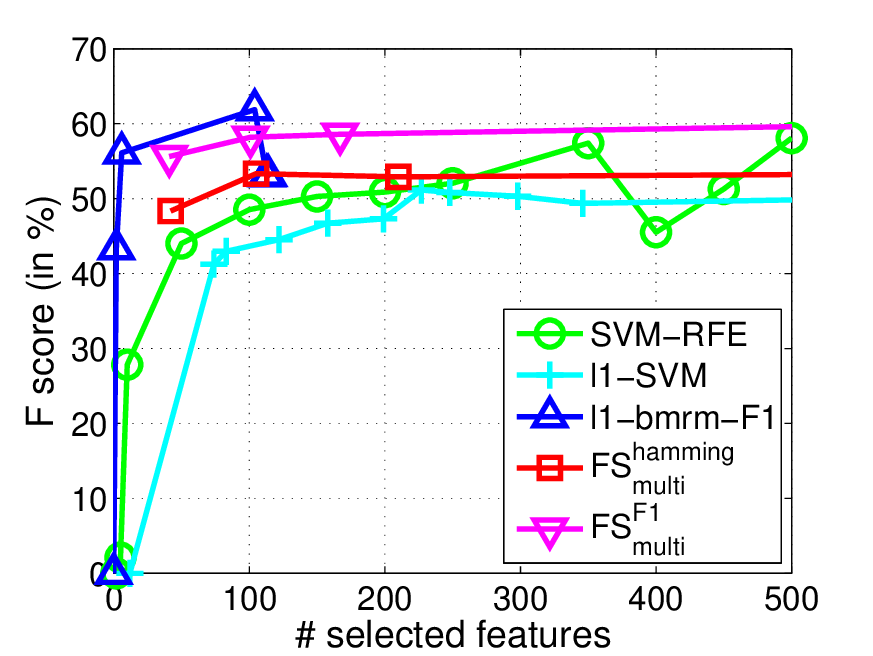}\\
(a) Desert & (b) Mountains \\
\includegraphics[width=0.5\textwidth]{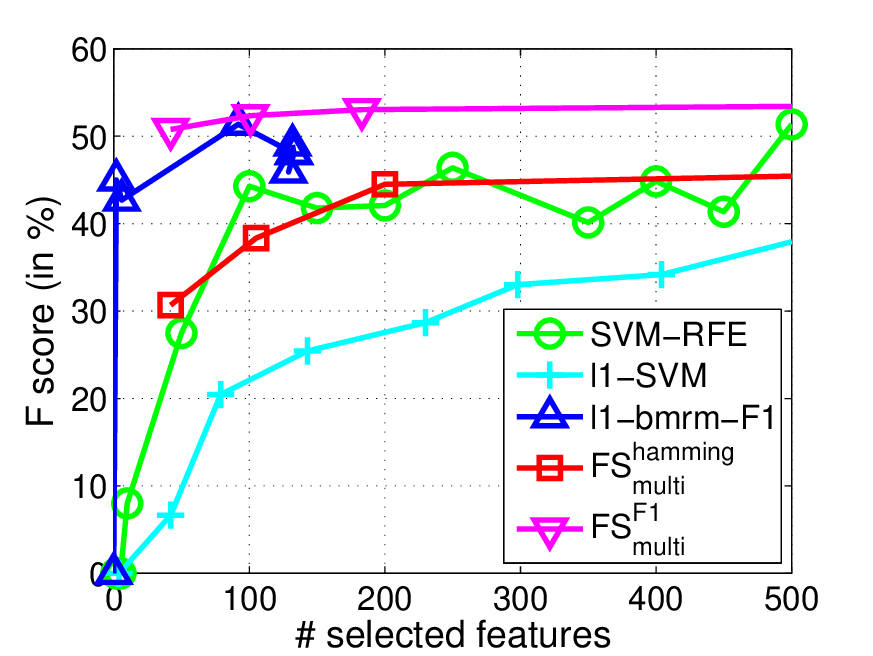}\!\!\!\!\!\!&\!\!\!\!\!\!
\includegraphics[width=0.5\textwidth]{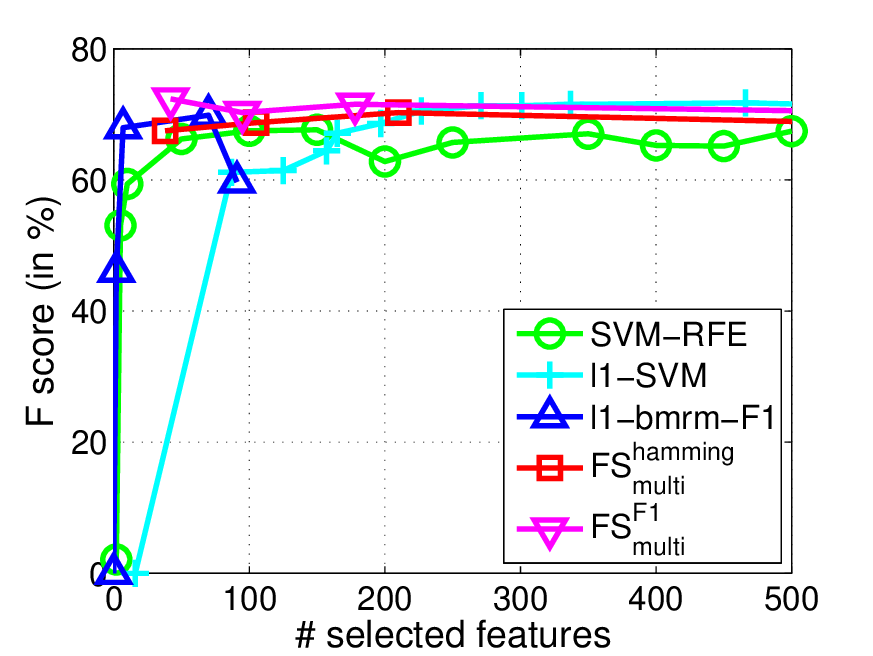}\\
(c) Sea & (d) Sunset \\
\multicolumn{2}{c}{\includegraphics[width=0.5\textwidth]{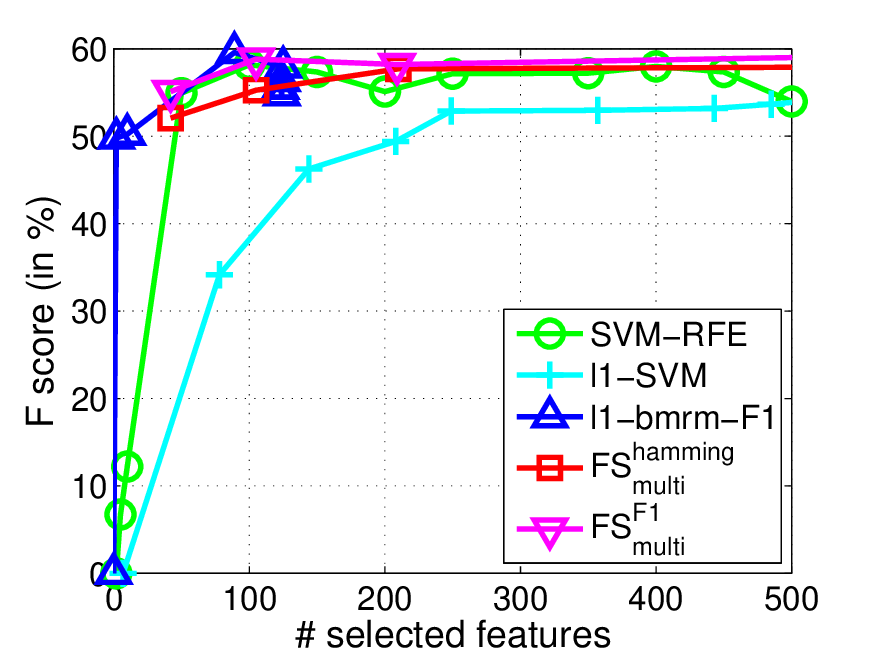}}\\
\multicolumn{2}{c}{(e) Trees}\\
\end{tabular}
 \caption{ Testing $F_1$ scores on \emph{Image} dataset. }\label{fig:dense-data}
\end{figure}

\begin{table*}
\caption{The macro-average testing performance comparisons among different methods. The quantities in the parentheses represent won/lost of the current method comparing with $\textrm{FS}_{multi}^{\Delta} $. The last column indicates the average number of features is actually used in the current method for a specific measure. The symbol '*' indicates the level of significance at $0.95$ according to t-test applied to pairs of results over classes} \label{tab:average_performance}
\begin{center}
\begin{tabular}{l|l|ccc|c}
\hline
Dataset  & method & $F_1$  & $Rec@2p$ & $PRBEP$ & \#selected features \\
\hline
\hline
 & $\textrm{FS}_{multi}^{\Delta} $ & 92.07 & 95.77 & 93.25 & 787.6/658.9/508.3\\
Sector & $\textrm{FS}_{multi}^{hamming} $& 84.99 $(12/91)^*$& 90.01 $(0/71)^*$& 85.54 $(0/86)^*$ & 689.2\\
 & $\textrm{SVM}_{multi}^{\Delta} $ & 33.35 $(1/104)^*$& 95.52 $(11/19)$& 91.24 $(11/47)^*$ & 55,197\\
\hline
\hline
 & $\textrm{FS}_{multi}^{\Delta} $ & 77.56 & 91.21 & 81.46 & 1,301 / 1,186 / 931\\
News20 & $\textrm{FS}_{multi}^{hamming} $& 49.61 $(0/20)^*$  & 66.32 $(0/20)^*$ & 52.14 $(0/20)^*$ &485.1\\
 & $\textrm{SVM}_{multi}^{\Delta} $ & 55.53 $(0/20)^*$  & 93.08 (16/2) & 80.83 (6/11) & 62,061\\
\hline
\end{tabular}
\end{center}
\end{table*}

In this subsection, we demonstrate the specific multivariate performance measures are important to select features for real applications. In particular, we  evaluate $F_1$ measure (commonly used performance measure) for the task of image retrieval. Due to the success of transforming multiple instance learning into a feature selection problem by embedded instance selection,  we use the same strategy in Algorithm 4.1 of \cite{Chen06} to construct a dense and high-dimensional dataset on a preprocessed image data~\footnote{http://lamda.nju.edu.cn/data\_MIMLimage.ashx}.
This dataset  is used in \cite{Zhou07} for multi-instance learning. It
contains five categories and $2,000$ images. Each image is represented as a bag of nine instances generated by the SBN method \cite{Maron98}. Each image bag is represented by a collection of nine 15-dimensional feature vectors.
After that, following \cite{Chen06}, the natural scene image retrieval problem turns out to be a feature selection task to select relevant embedded instances for prediction. The \emph{Image}  dataset are split randomly with the proportion of 60\% for training and 40\% for testing (Table \ref{tab:datasets}). Since $F_1$-score is used for performance metric, we perform $\textrm{FS}_{multi}^{\Delta}$
for $F_1$-score, namely $\textrm{FS}_{multi}^{F_1}$ as well as other state-of-the-art feature selection methods. As mentioned above, FGM and $\textrm{FS}_{multi}^{hamming}$ have similar performances, we will not report the results of FGM here.  $\textrm{FS}_{multi}^{hamming}$ and $\textrm{FS}_{multi}^{\Delta}$ use the fixed $C=10\times n$. For other methods, we use the previous settings.  The testing $F_1$ values of all  methods on each category are reported in Figure \ref{fig:dense-data}.

From Figure \ref{fig:dense-data}, we observe that  $\textrm{FS}_{multi}^{F_1}$ and $\textrm{FS}_{multi}^{hamming}$ achieve significantly improved performance over $l_1$-SVM in term of $F_1$-score especially when choosing less than $100$ features. Moreover, SVM-RFE also outperforms $l_1$-SVM  on three categories out of five.
This verifies that
$\ell_1$ penalty does not perform as well as $\ell_0$ methods like $\textrm{FS}_{multi}^{F_1}$ and $\textrm{FS}_{multi}^{hamming}$ on dense and high-dimensional datasets. It is possibly because $\ell_1$-norm penalty is very sensitive to dense and noisy features.
We also observe that $\textrm{FS}_{multi}^{F_1}$ performs better than $\textrm{FS}_{multi}^{hamming}$ and SVM-RFE on four over five categories. $l_1$-bmrm-F$_1$ performs competitively but it is unstable and time-consuming as shown in Section \ref{sec:time}. All these facts imply that directly optimizing $F_1$ measure is useful to boost $F_1$ performance measure, and our proposed $\textrm{FS}_{multi}^{F_1}$ is efficient and effective.

\subsection{Multivariate Performance Measures for Document Retrieval}

In this subsection, we focus on feature selection for different multivariate performance measures on imbalanced text data shown in Table \ref{tab:datasets}. For multiclass classification problems,  one vs. rest strategy is used. The comparing model is $\textrm{SVM}^{perf}$ \footnote{www.cs.cornell.edu/People/tj/svm\_light/svm\_perf.html}. Following \cite{Joachims05}, we use the same notation $\textrm{SVM}_{multi}^{\Delta}$  for different multivariate performance measures. The command used for training $\textrm{SVM}^{perf}$ can work for different measures by -$l$ option \footnote{ svm\_perf\_learn -c $C_{perf}$ -w 3 --b 0 train\_file train\_model}.
In our experiments, we search the $C_{perf}$ in the same range $[2^{-6},\ldots,2^6]$   as in  \cite{Joachims05}.  We choose the one which demonstrates the best performance of $\textrm{SVM}_{multi}^{\Delta}$  to each multivariate performance measure for comparison. $\textrm{FS}_{multi}^{\Delta}$ and $\textrm{FS}_{multi}^{hamming}$ fix $C_{FGM_{multi}} = 0.1 \times n$ for \emph{News20} except $5.0 \times n$ for \emph{Sector}.
For $Rec@k$, we use $k$ as twice the number of positive examples, namely \emph{Rec@2p}. The evaluation for this measure uses the same strategy to label twice the number of positive examples as positive in the test datasets, and then calculate \emph{Rec@2p}.

\begin{figure}
\centering
\begin{tabular}{cc}
\includegraphics[width=0.5\textwidth]{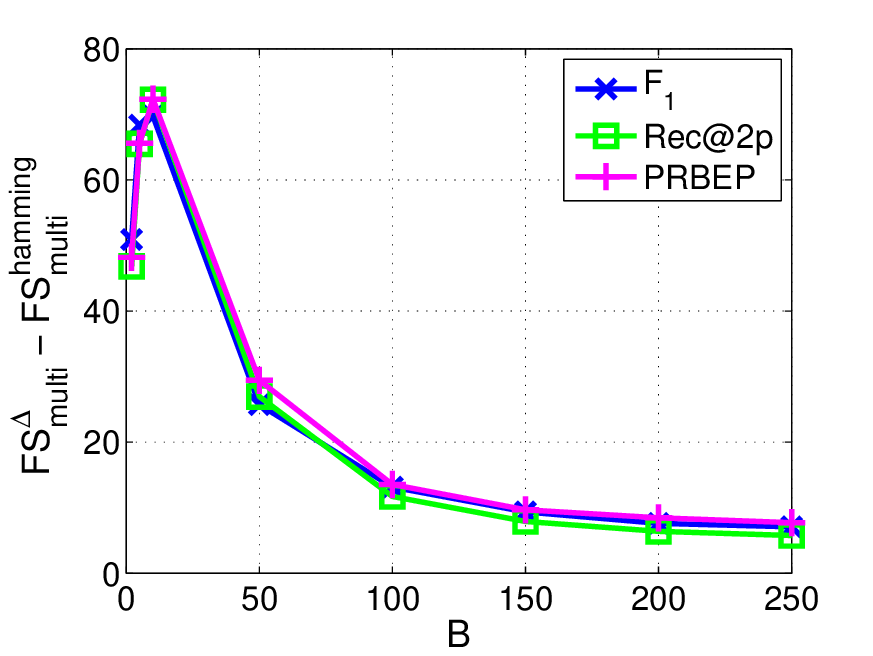} \!\!\!\!\!\!&\!\!\!\!\!\!
\includegraphics[width=0.5\textwidth]{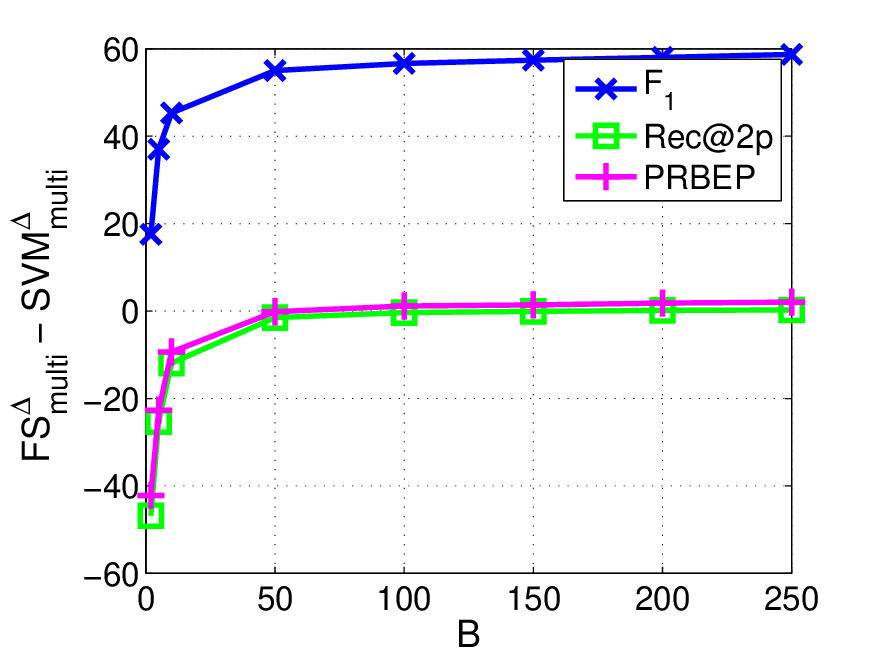}\\
\multicolumn{2}{c}{(a) Sector}\\
\includegraphics[width=0.5\textwidth]{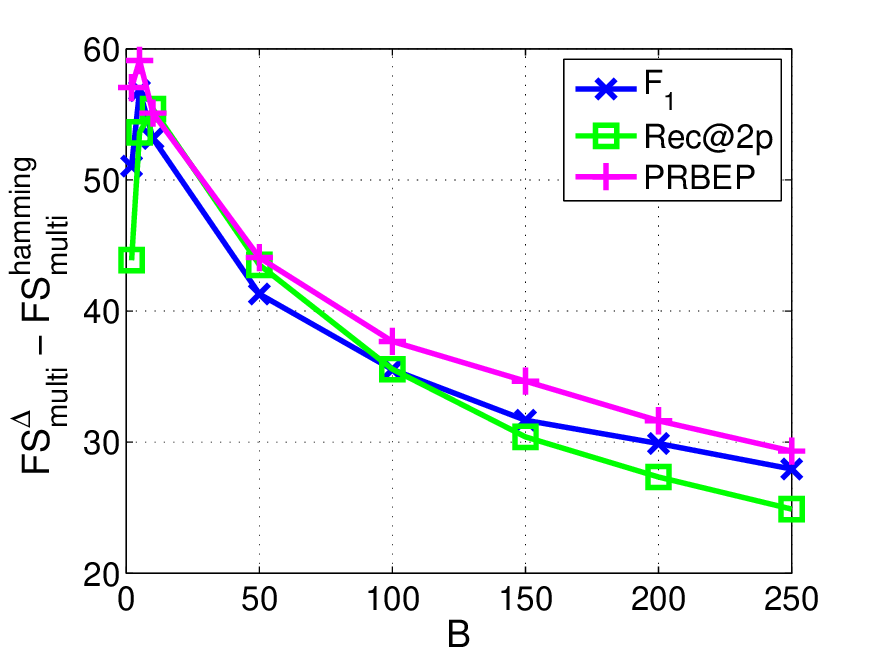}\!\!\!\!\!\!&\!\!\!\!\!\!
\includegraphics[width=0.5\textwidth]{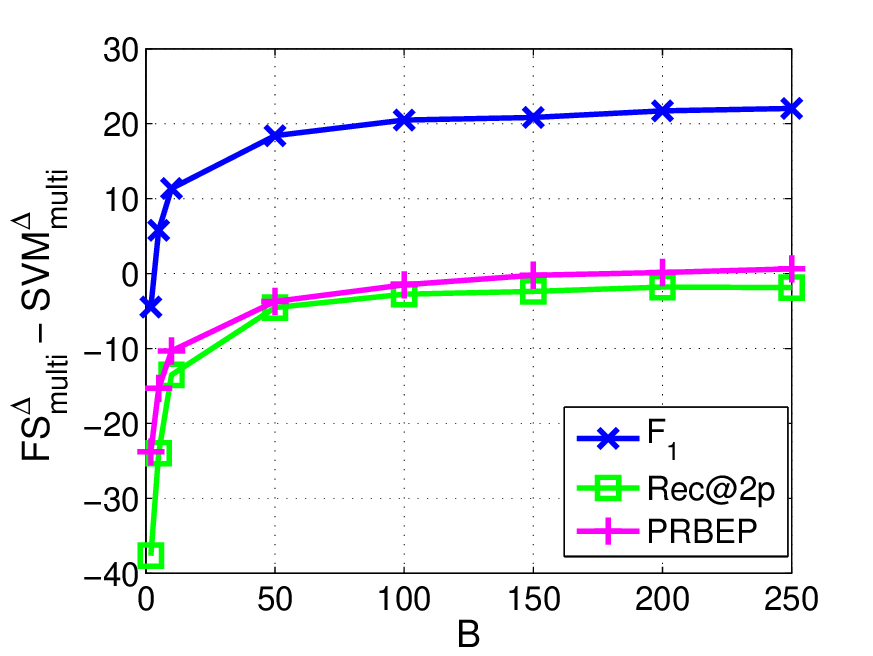}\\
\multicolumn{2}{c}{(b) News20}\\
\end{tabular}
\caption{ The average performance improvement of $\textrm{FS}_{multi}^{\Delta}$ with varying $B$ on different datasets. }\label{fig:average_performance}
\end{figure}

Table \ref{tab:average_performance} shows the macro-average of the performance over all classes in a collection in which both $\textrm{FS}_{multi}^{\Delta}$ and $\textrm{FS}_{multi}^{hamming}$ at $B=250$ are listed.
The improvement of $\textrm{FS}_{multi}^{\Delta}$ over $\textrm{FS}_{multi}^{hamming}$ and $\textrm{SVM}_{multi}^{\Delta}$
 with respect to different  $B$ values are  reported in Figure \ref{fig:average_performance}. From Table \ref{tab:average_performance},
$\textrm{FS}_{multi}^{\Delta}$ is consistently better than $\textrm{FS}_{multi}^{hamming}$ on all multivariate performance measures and two multiclass datasets. Similar results can be obtained comparing with $\textrm{SVM}_{multi}^{\Delta}$, while the only exception is the measure \emph{Rec@2p} on \emph{News20} where $\textrm{SVM}_{multi}^{\Delta}$ is a little better than $\textrm{FS}_{multi}^{\Delta}$. The largest gains are observed for $F_1$ score on all two text classification tasks. This implies that a small number of features selected by $\textrm{FS}_{multi}^{\Delta}$ is enough to obtain comparable or even better performances for different measures than $\textrm{SVM}_{multi}^{\Delta}$ using all features.

\begin{figure}
\centering
\begin{tabular}{cc}
\includegraphics[width=0.5\textwidth]{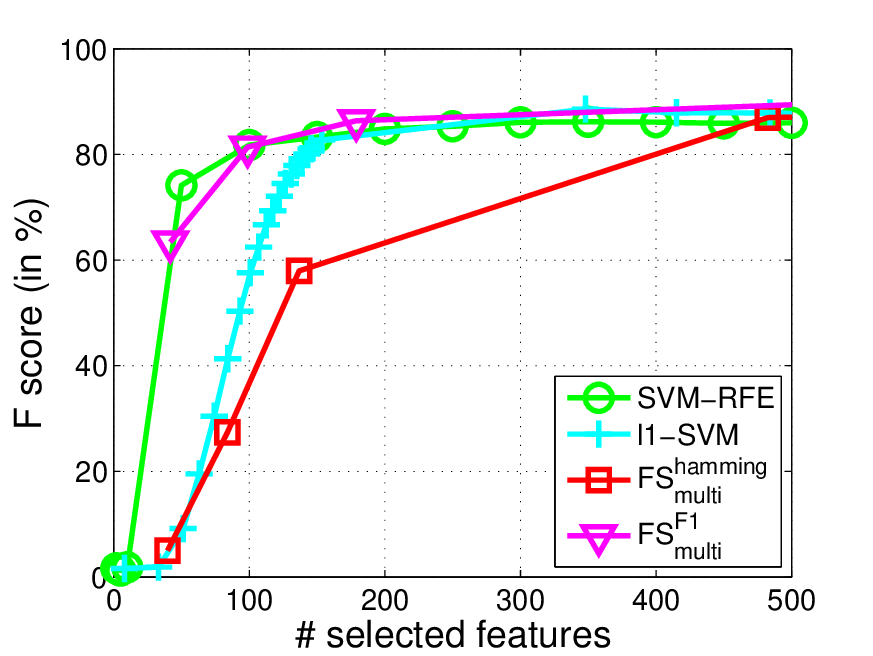}\!\!\!\!\!\!&\!\!\!\!\!\!
\includegraphics[width=0.5\textwidth]{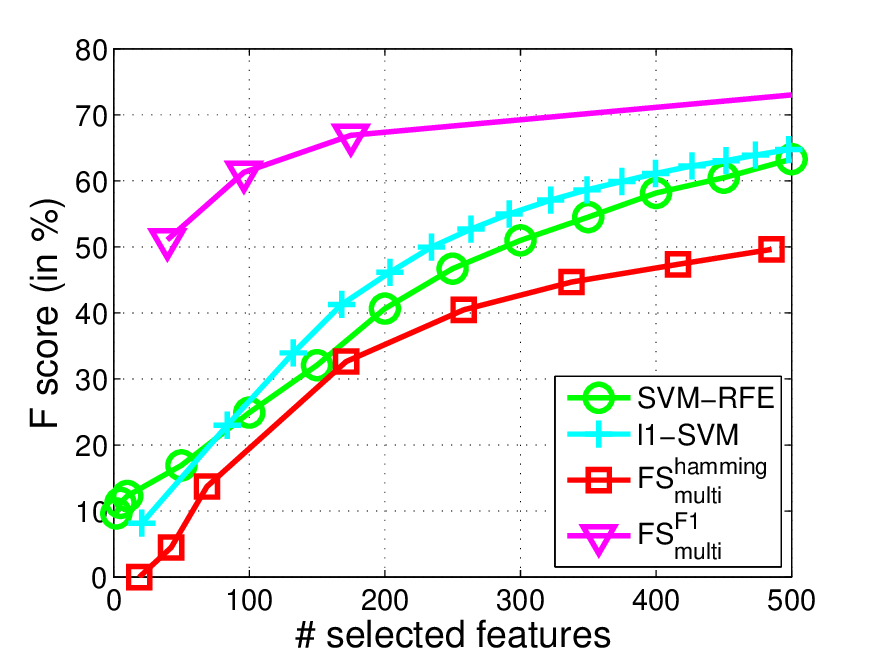}\\
(a) Sector & (b) News20  \\
\end{tabular}
\caption{ Testing $F_1$ in terms of the average number of selected features on Sector and News20. } \label{fig:average_f1}
\end{figure}

From Figure \ref{fig:average_performance}, $\textrm{FS}_{multi}^{\Delta}$ consistently performs better than $\textrm{FS}_{multi}^{hamming}$ for all of the multivariate performance measures from the figures in the left-hand side. Moreover, the figures in the right-hand side show that the small number of features are good for $F_1$ measures, but poor for other measures. As the number of features increases, $Rec@2p$ and \emph{PRBEP} can approach to the results of $\textrm{SVM}_{multi}^{\Delta}$ and all curves become flat. The performance of $PRBEP$ and $Rec@2p$ is relatively stable when sufficient features are selected, but our method can choose very few features for fast prediction. For $F_1$ measure, our method is consistently better than $\textrm{SVM}_{multi}^{\Delta}$, and the results show significant improvement over all range of $B$. This improvement may be due to the reduction of noisy or non-informative features. Furthermore, $\textrm{FS}_{multi}^{\Delta}$ can achieve better performance measures than $\textrm{FS}_{multi}^{hamming}$.

We also compared different feature selection algorithms such as SVM-RFE and $l_1$-SVM on \emph{Sector} and \emph{News20} in the same setting as the previous sections. The results in terms of F1 measure are reported in Figure \ref{fig:average_f1}. We clearly observe that $\textrm{FS}_{multi}^{\Delta}$ outperforms $l_1$-SVM on both datasets, and comparable or even better than SVM-RFE. For a small number of features, $\textrm{FS}_{multi}^{\Delta}$ can still demonstrate very good F1 measure.

\section{Conclusion} \label{sec:conclusion}
In this paper, we propose a generalized sparse regularizer for feature selection, and the unified feature selection framework for general loss functions. We particularly study in details for multivariate losses. To solve the resultant optimization problem, a two-layer cutting plane algorithm was proposed. The convergence property of the proposed algorithm is studied.
Moreover, connections to a variety of state-of-the-art feature selection methods are discussed in details.
 A variety of analyses by comparing with the various feature selection methods show that the proposed method is superior to others. Experimental results show that the proposed method is comparable with FGM and SVM-RFE and better than $l_1$ models on feature selection task, and outperforms SVM for multivariate performance measures on full set of features.

\section*{Acknowledgements}

This work was supported by Singapore A*star under Grant SERC 112 280 4005

\appendices

\section{Proof of Proposition 1}

Since the loss term $\Delta(\overline{y}',\overline{y}')=0$ for all $\overline{y}' \in \mathcal{Y}$, we can equivalently transform Problem
\begin{eqnarray*}
\min_{\textbf{w}_1,\ldots,\textbf{w}_T,\xi \geq 0} \!\!\!\!\!&&\!\!\!\!\! \frac{1}{2} \left( \sum_{t=1}^T \|\textbf{w}_t\|_2 \right)^2 + C \xi \label{prob:mkl-primal}\\
\textrm{s.t.} \!\!\!\!\!&&\!\!\!\!\! \xi \geq  b_{\overline{y}'} - \sum_{t=1}^T \langle \textbf{w}_t, \textbf{a}_{\overline{y}'}^t \rangle, \forall \overline{y}' \in \overline{\mathcal{Y}} \backslash \overline{y} \nonumber,
\end{eqnarray*}
into the following optimization problem
\begin{eqnarray*}
\min_{\textbf{w}_1,\ldots,\textbf{w}_T,\xi \geq 0} \!\!\!\!\!&&\!\!\!\!\! \frac{1}{2} \left( \sum_{t=1}^T \|\textbf{w}_t\|_2 \right)^2 + C \xi\\
\textrm{s.t.} \!\!\!\!\!&&\!\!\!\!\! \xi \geq  b_{\overline{y}'} - \sum_{t=1}^T \langle \textbf{w}_t, \textbf{a}_{\overline{y}'}^t \rangle, \forall \overline{y}' \in \overline{\mathcal{Y}} .
\end{eqnarray*}
By introducing a new variable $u \in \mathbb{R}$ and moving out summation operator from objective to be a constraint, we can obtain the equivalent optimization problem as
\begin{eqnarray*}
\min_{\textbf{w}, \xi \geq 0} && \frac{1}{2} u^2 + C \xi \\
\textrm{s.t.} && \xi \geq b_{\overline{y}'} - \sum_{t=1}^T \langle \textbf{w}_t, \textbf{a}_{\overline{y}'}^t \rangle, \forall \overline{y}' \in \overline{\mathcal{Y}} \\
              && \sum_{t=1}^T \|\textbf{w}_t\| \leq u.
\end{eqnarray*}
We can further simplify above problem by introducing another variables $\rho \in \mathbb{R}^m$ such that $\|\textbf{w}_t\| \leq \rho_t$, $\forall t=1,\ldots,T$, to be
\begin{eqnarray*}
\min_{\textbf{w}, u, \rho, \xi \geq 0} && \frac{1}{2} u^2 + C \xi \\
\textrm{s.t.} && \xi \geq b_{\overline{y}'} - \sum_{t=1}^T \langle \textbf{w}_t, \textbf{a}_{\overline{y}'}^t \rangle, \forall \overline{y}' \in \overline{\mathcal{Y}} \\
              && \sum_{t=1}^T \rho_t \leq u \\
              && ||\textbf{w}_t|| \leq \rho_t, \forall t=1,\ldots,T.
\end{eqnarray*}
We know that for each $t$, $\|\textbf{w}_t\| \leq \rho_t$ is a second-order cone constraint. Following the recipe of \cite{Boyd04}, the self-dual cone $\|\textbf{v}_t\|_2 \leq \eta_t, \forall t=1,\ldots,T$ can be introduced to form the Lagrangian function as follows
\begin{eqnarray*}
&&\mathcal{L}(\textbf{w},\xi,u,\rho; \alpha,\tau,\gamma,\textbf{v},\eta) \\
&=& \frac{1}{2} u^2 + C \xi - \sum_{\overline{y}'} \alpha_{\overline{y}'} \left(\xi - b_{\overline{y}'} +  \sum_{t=1}^T \langle \textbf{w}_t, \textbf{a}_{\overline{y}'}^t \rangle \right)- \tau \xi  \\
&&+ \gamma \Bigg( \sum_{t=1}^T \rho_t - u \Bigg)- \sum_{t=1}^T (\langle \textbf{v}_t, \textbf{w}_t \rangle + \eta_t \rho_t),
\end{eqnarray*}
with dual variables $\alpha_t \in \mathbb{R}_+$, $\tau \in \mathbb{R}_+$, $\gamma \in \mathbb{R}_+$. The derivatives of the Lagrangian with respect to the primal variables have to vanish which leads to the following KKT conditions:
\begin{eqnarray*}
&&\textbf{v}_t = - \sum_{\overline{y}'} \alpha_{\overline{y}'} \textbf{a}_{\overline{y}'}^t, \forall t=1,\ldots,T \\
&&C - \sum_{\overline{y}'} \alpha_{\overline{y}'} - \tau = 0\\
&&u = \gamma \\
&&\gamma = \eta_t, \forall t=1,\ldots,T
\end{eqnarray*}
By substituting all the primal variables with dual variables by above KKT conditions, we can obtain the following dual problem,
\begin{eqnarray*}
\max_{\alpha,\gamma} && -\frac{1}{2} \gamma^2 + \sum_{\overline{y}'} \alpha_{\overline{y}'} b_{\overline{y}'}  \\
\textrm{s.t.} && \Big\| \sum_{\overline{y}'} \alpha_{\overline{y}'} \textbf{a}_{\overline{y}'}^t \Big\| \leq \gamma, \forall t=1,\ldots,T\\
              &&\sum_{\overline{y}'} \alpha_{\overline{y}'} \leq C, ~\alpha_{\overline{y}'} \geq 0, \forall \overline{y}' \in \overline{\mathcal{Y}}
\end{eqnarray*}
By setting $\theta = \frac{1}{2}\gamma^2$ and $\mathcal{A}=\{\sum_{\overline{y}'} \alpha_{\overline{y}'} \leq C, \alpha_{\overline{y}'} \geq 0, \forall \overline{y}' \in \overline{\mathcal{Y}}\}$, we can reformulate above problem as
\begin{eqnarray*}
\max_{\theta, \alpha \in \mathcal{A}} && -\theta + \sum_{\overline{y}'} \alpha_{\overline{y}'} b_{\overline{y}'}  \\
\textrm{s.t.} && \frac{1}{2} \alpha^T Q^t \alpha \leq \theta, \forall t=1,\ldots,T
\end{eqnarray*}
where $Q_{\overline{y}',\overline{y}''}^t = \langle \textbf{a}_{\overline{y}'}^t, \textbf{a}_{\overline{y}''}^t \rangle$. According to the property of self-dual cone \cite{Bach04}, we can obtain the primal solution from its dual as $\textbf{w}_t = - \mu_t \textbf{v}_t = \mu_t \sum_{\overline{y}'} \alpha_{\overline{y}'} \textbf{a}^t_{\overline{y}'}$ where $\mu_j$ is the dual variable of the $j^{th}$ quadratic constraint such that $\sum_{j=1}^m \mu_j = 1,\mu_j \in \mathbb{R}_+, \forall j=1,\ldots,m$. By constructing Lagrangian with dual variables $\mu$ with respect to $\theta$, we can recover Problem
\begin{equation}
\max_{\alpha \in \mathcal{A}}\min_{\mu \in \mathcal{M}_T} -\frac{1}{2} \sum_{\overline{y}'} \sum_{\overline{y}''}  \alpha_{\overline{y}'}  \alpha_{\overline{y}''} \left( \sum_{t=1}^T \mu_t Q_{\overline{y}',\overline{y}''}^{\textbf{d}^t} \right) +  \sum_{\overline{y}'} \alpha_{\overline{y}'} b_{\overline{y}'}, \label{prob:svm-mpm-ours_dual_mu}
\end{equation}
where $\mathcal{M}_T = \{ \sum_{t=1}^T \mu_t = 1, \mu_t \geq 0, \forall t=1,\ldots,T\}$. This completes the proof.

\section{Proof of Theorem 2}
Given the Problem
\begin{eqnarray}\label{eq:minimax}
\min_{\alpha \in \mathcal{A}} \max_{\textbf{d} \in \mathcal{D}}\mathcal{F}_\textbf{d}(\alpha) \!&\!\!\mbox{or}\!\!&\! \min_{\alpha \in \mathcal{A},\gamma}\gamma \;:\; \gamma\geq \mathcal{F}_\textbf{d}(\alpha),\; \forall\textbf{d} \in \mathcal{D},
\end{eqnarray}
 we have the equivalent optimization problem as
\begin{eqnarray*}
\max_{\alpha \in \mathcal{A}, \gamma} && - \gamma \\
\textrm{s.t.} && \gamma \geq \mathcal{F}_{\mathbf{d}} (\alpha), \forall \mathbf{d} \in \mathcal{D}.
\end{eqnarray*}
The outer layer of Algorithm 2 can generate a sequence of configurations of $\mathbf{d}$ as $\{\mathbf{d}^1,\ldots,\mathbf{d}^k\}$ after $k$ iterations. In the $k$th iteration, the most violated constraint $d^{k+1}$ is found in terms of $\alpha_k$, so that $\mathcal{F}_{\mathbf{d}^{k+1}} (\alpha_k) = \max_{\mathbf{d} \in \mathcal{D}} \mathcal{F}_{\mathbf{d}} (\alpha)$ according to Problem $\textbf{d}^t = \arg \max_{\textbf{d} \in \mathcal{D}} \mathcal{F}_{\textbf{d}}(\alpha^t)$. Hence, we can construct two sequences $\{\underline{\gamma}_k\}$ and $\{\overline{\gamma}_k\}$ such that
\begin{eqnarray}
\underline{\gamma}_k &=& \max_{1 \leq t \leq k} \mathcal{F}_{\mathbf{d}^t} (\alpha_t) \label{eq:lower-bound}\\
\overline{\gamma}_k &=& \min_{1 \leq t \leq k} \mathcal{F}_{\mathbf{d}^{t+1}} (\alpha_t) =  \min_{1 \leq t \leq k} \max_{\mathbf{d} \in \mathcal{D}} \mathcal{F}_{\mathbf{d}} (\alpha_t)
\end{eqnarray}
Suppose that we can solve $\min_{\alpha \in \mathcal{A}} \max_{1 \leq t \leq k} \mathbf{F}_{\mathbf{d}^t} (\alpha)$ exactly. Due to the equivalence to Problem (\ref{prob:svm-mpm-ours_dual_mu}), it means that we can obtain the exact solution of the problem (\ref{prob:svm-mpm-ours_dual_mu}). Based on this assumption, equation (\ref{eq:lower-bound}) can be further reformed as
\begin{eqnarray}
\underline{\gamma}_k = \max_{1 \leq t \leq k} \mathcal{F}_{\mathbf{d}^t} (\alpha_t) = \min_{\alpha \in \mathcal{A}} \max_{1 \leq t \leq k}  \mathcal{F}_{\mathbf{d}^t} (\alpha_t).
\end{eqnarray}
This turns out to be the same problem of FGM \cite{Tan10}. For self-completeness, we give the theorem as follows,
\begin{theorem}[\cite{Tan10}]
Let $(\alpha^*,\gamma^*)$ be the globally optimal solution pair of Problem (\ref{eq:minimax}), sequences $\{\underline{\gamma}_k\}$ and $\{\overline{\gamma}_k\}$ have the following property
\begin{equation}
\underline{\gamma}_k \leq \gamma_k \leq \overline{\gamma}_k. \label{eq:upper-lower-bound}
\end{equation}
As $k$ increases, $\{\underline{\gamma}_k\}$ is monotonically increasing and $\{\overline{\gamma}_k\}$ is monotonically decreasing.
\end{theorem}
Based on above theorem, global optimal solution can be obtained after a finite number of iterations. However, the assumption of the accurate solution for (\ref{prob:svm-mpm-ours_dual_mu}) usually has no formal guarantee. We have already proven in Theorem 1 that the inner problem of Algorithm 2 can reach the desired precision $\epsilon$ after a finite number of iterations by Algorithm 1. Therefore, according to Algorithm 2, we can construct the following sequence
\begin{eqnarray}
\underline{\gamma}_k' = \max_{1 \leq t \leq k} \mathcal{F}_{\mathbf{d}^t} (\alpha_t) \leq \min_{\alpha \in \mathcal{A}} \max_{1 \leq t \leq k}  \mathcal{F}_{\mathbf{d}^t} (\alpha_t) + \epsilon. \label{eq:our-lower-bound}
\end{eqnarray}
By combining inequalities (\ref{eq:upper-lower-bound}) and (\ref{eq:our-lower-bound}), we obtain the following inequalities
\begin{equation}
\underline{\gamma}_k' - \epsilon \leq \underline{\gamma}_k \leq \gamma_k \leq \overline{\gamma}_k.
\end{equation}
After a finite number of iterations, the global optimal solution is $\gamma^* = \underline{\gamma}_k = \gamma_k = \overline{\gamma}_k$. Hence, the solution of the Algorithm 2 may be not less than the lower bound $\underline{\gamma}_k'$ by $\epsilon$. It is complete for Theorem 2.

\bibliographystyle{plain}
\bibliography{group_feature_selection}

\end{document}